\documentclass[10pt,journal,compsoc]{IEEEtran}
\ifCLASSOPTIONcompsoc
  \usepackage[nocompress]{cite}
\else
  \usepackage{cite}
\fi
\ifCLASSINFOpdf
\else
\fi

\usepackage{times}
\usepackage{epsfig}
\usepackage{amsmath}
\usepackage{amssymb}
\usepackage{dsfont}
\usepackage{xcolor}
\usepackage{graphicx}
\usepackage{bm}
\usepackage{enumitem}
\usepackage[utf8]{inputenc}
\usepackage{multirow}
\DeclareUnicodeCharacter{2212}{-}

\begin{document}
%
\title{I2F: A Unified Image-to-Feature Approach for Domain Adaptive Semantic Segmentation}
%
%
%
%

\author{\IEEEauthorblockN{Haoyu~Ma\textsuperscript{\textsection}, Xiangru~Lin\textsuperscript{\textsection} and~Yizhou~Yu,~\IEEEmembership{Fellow,~IEEE}}
\IEEEcompsocitemizethanks{
\IEEEcompsocthanksitem This work was supported in part by Hong Kong Research Grants Council through Research Impact Fund (Grant R-5001-18). H. Ma was supported by the Hong Kong PhD Fellowship. (Corresponding author: Yizhou Yu).
\IEEEcompsocthanksitem  H. Ma, X. Lin and Y. Yu are with the Department of Computer Science, the University of Hong Kong, Pokfulam Road, Hong Kong. E-mail: mahaoyu@connect.hku.hk; xrlin2@cs.hku.hk; yizhouy@acm.org.
\IEEEcompsocthanksitem \textsuperscript{\textsection} H. Ma and X. Lin have equal contribution.
}
}
%
%

\markboth{IEEE Transactions on Pattern Analysis and Machine Intelligence}%
{Ma \MakeLowercase{\textit{et al.}}: A Unified Image-to-Feature Approach for Domain Adaptive Semantic Segmentation}
%



\IEEEtitleabstractindextext{
\begin{abstract}
Unsupervised domain adaptation (UDA) for semantic segmentation is a promising task freeing people from heavy annotation work. However, domain discrepancies in low-level image statistics and high-level contexts compromise the segmentation performance over the target domain. A key idea to tackle this problem is to perform both image-level and feature-level adaptation jointly. Unfortunately, there is a lack of such unified approaches for UDA tasks in the existing literature. This paper proposes a novel UDA pipeline for semantic segmentation that unifies image-level and feature-level adaptation. Concretely, for image-level domain shifts, we propose a global photometric alignment module and a global texture alignment module that align images in the source and target domains in terms of image-level properties. For feature-level domain shifts, we perform global manifold alignment by projecting pixel features from both domains onto the feature manifold of the source domain; and we further regularize category centers in the source domain through a category-oriented triplet loss, and perform target domain consistency regularization over augmented target domain images. Experimental results demonstrate that our pipeline significantly outperforms previous methods. In the commonly tested GTA5$\rightarrow$Cityscapes task, our proposed method using Deeplab V3+ as the backbone surpasses previous SOTA by 8\%, achieving 58.2\% in mIoU.
\end{abstract}

}

\maketitle

\IEEEdisplaynontitleabstractindextext

%
\IEEEpeerreviewmaketitle

\IEEEraisesectionheading{\section{Introduction}\label{sec:introduction}}

%
%
%
%
\IEEEPARstart{S}{emantic} segmentation, a classical and fundamental research task in computer vision, aims to assign category labels to individual pixels in an image. It has been extensively investigated and has inspired many downstream applications including autonomous driving~\cite{autodrive1,autodrive2} and medical image analysis~\cite{medical1,medical2,medical3}. Although the performance of existing semantic segmentation models have enjoyed a significant improvement in the wave of deep neural networks~\cite{deeplab,oldfcn,crf}, training a semantic segmentation model usually requires a large number of images with pixel-level annotations, the collection process of which is laborious and time-consuming. Unsupervised Domain Adaptation (UDA) for semantic segmentation is an alternative to avoid the data annotation problem:
it aims at learning a well-performing model from an unlabeled target dataset by jointly exploiting labeled images from a different source dataset (the label spaces of the two datasets must be compatible). However, domain shifts/discrepancies exist between different datasets. The most obvious differences are low-level image statistics related to colors, textures, or even illumination conditions. These differences can be partly alleviated by image-level adaptation. However, there are also object-level differences, such as object poses and spatial distributions, between different datasets, which give rise to different feature distributions. All these domain shifts have a detrimental impact on the final performance of the semantic segmentation model. Therefore, it is crucial to learn a feature representation capable of overcoming both image-level and feature-level domain shifts for unsupervised domain adaptive semantic segmentation.

The causes of domain shifts/discrepancies have been extensively studied in previous works. In general, the primary causes can be categorized into image-level domain shifts and feature-level domain shifts. Image-level domain shifts refer to the differences in imaging conditions, such as lighting and settings in the camera imaging pipeline. They affect the overall appearance of an image and have a subtle influence on feature-level distributions. Existing work addressing image-level domain shifts is in general based on image-level style transfer, which makes use of deep models such as generative models or image-to-image translation models~\cite{cgan, cgan2}, or Fourier Transformation~\cite{FDA}. We term these methods image-level adaptation methods. These methods have proven that transferring image styles or aligning feature distributions can bring the two domains closer. However, generative methods usually require a computationally expensive training process, whose instability is notorious. Generative models also suffer from mode collapse, which makes the range of the generated features unusually small (more explanation in Related Work). On the other hand, the Fourier Transformation based method~\cite{FDA} produces inferior style-transferred images, as shown in Figure~\ref{fig:qualitative_gpa}.

We have observed that previous work in domain adaptive semantic segmentation focusing on image-level domain alignment~\cite{dagan3,cgan2} usually has inferior final segmentation performance in comparison to recent work that adopts a more complete pipeline~\cite{intra_domain,BDL}. Such recent work further demonstrates that replacing the original source domain images with image-level domain aligned images can further improve the final performance of feature-level adaptation techniques.
This indicates that the domain gap can only be partially alleviated with aforementioned image-level adaptation methods, and feature-level alignment can still benefit from an extra image translation module. Therefore, feature-level adaptation is still necessary after image-level adaptation.
For feature-level adaptation, a common practice in previous studies employs an adversarial method~\cite{BDL,fgan}, which considers features from the source and target domains aligned if they cannot be distinguished by a trained discriminator. But adversarial methods tend to generate a narrow range of feature distributions to fool the discriminator. When different images share similar feature distributions, trained models would have poor generalization performance. On the other hand, to perform category-level feature adaptation, some existing methods use category anchors computed in the source domain to align the two domains~\cite{CAG,stuff_things}, which can be regarded as imposing hard constraints on category-level feature distributions. This method ignores feature distances across different categories, and categories with similar feature distributions in the source domain may still have similar ones in the target domain, resulting in erroneous pseudo-labels when no supervision signals are available in the target domain. Our experiments demonstrate that imposing soft regularization on category-level feature distributions by adjusting the relative magnitude of inter-category and intra-category feature distances can improve model capacity.

According to the above analysis, performing either image-level adaptation or feature-level adaptation alone could not address domain shifts adequately. Moreover, existing work on UDA for semantic segmentation lacks a unified approach to minimize domain shifts. Therefore, we approach the problem from both perspectives and propose a novel and efficient pipeline that unifies image-level and feature-level adaptation. For image-level domain shifts, we propose two novel and training-free image-level operation, called global photometric alignment and global texture alignment, to adapt images from the source domain to the target domain. However, image-level adaptation alone does not guarantee domain alignment in the feature space. Therefore, we devise a global manifold alignment module for feature-level adaptation. This module represents the source domain feature manifold with a set of atoms, and any pixel feature from the source domain or the target domain can be projected onto this manifold. By minimizing the projection errors between the input features and the manifold, all source and target domain features are aligned to the same manifold. To perform category-level feature adaptation, we also introduce two category-level feature distribution regularization methods: a category-oriented triplet loss is proposed in the source domain to softly regularize category centers by enlarging the margin between inter-category and intra-category feature distances. It is only adopted in the source domain because the measurement of inter-category and intra-category distances require reliable annotations that only exist in the source domain. The category-level feature adaptation method applied to the target domain is the self-supervised consistency regularization. This regularization makes the prediction on an augmented target image consistent with the pseudo-label of the corresponding non-augmented image, thus forcing the class labels of similar semantic contents to be consistent in the target domain. By addressing domain shifts from all perspectives simultaneously, experimental results demonstrate that our proposed method is capable of achieving significant performance improvements.

Domain adaptive semantic segmentation methods can be applied to either synthetic source images or real source images as long as there exist significant domain gaps. For the application to synthetic source images, we follow the common practice~\cite{CAG,c2f,BDL,FDA,fgan,intra_domain} and use the GTA5$\rightarrow$Cityscapes and SYNTHIA$\rightarrow$Cityscapes benchmarks to evaluate our proposed domain adaptation algorithm. In addition to synthetic source data, we also construct a new task on two open-source real-world endoscopic image datasets, Hyper-Kvasir~\cite{kvasir} and Piccolo~\cite{piccolo}. This task can serve as a new medical image benchmark for future studies in domain adaptive semantic segmentation. Experiment results on all three benchmarks demonstrate that our proposed method is capable of achieving significant performance improvements over existing state-of-the-art algorithms.

To this end, this paper is an extension of ~\cite{c2f} and the contributions of ~\cite{c2f} can be summarized as follows,
\begin{itemize}[noitemsep, nolistsep]
  \item A novel image-to-feature domain adaptive semantic segmentation pipeline is proposed to seamlessly combine coarse image-level adaptation with category-level feature distribution regularization.
  \item Two novel and effective category-level regularization methods are proposed to deal with the source and target domain shifts, respectively. The first one is category-oriented triplet loss which regularizes category centers in the source domain, and the second one performs target domain consistency regularization over augmented target domain images.
  \item The proposed method in~\cite{c2f} outperforms all previous methods, achieving state-of-the-art performances on both GTA5$\rightarrow$Cityscapes and SYNTHIA$\rightarrow$Cityscapes benchmarks.
\end{itemize}

Compared to the conference version~\cite{c2f}, this paper gives a more complete introduction and analysis of the proposed non-adversarial image-to-feature domain adaptive semantic segmentation pipeline. We provide more insights and discussions about the modules proposed in~\cite{c2f}. More importantly, we extend our work in ~\cite{c2f} by introducing global manifold alignment in the high-level feature space. This manifold alignment algorithm serves as a feature-level adaptation strategy complementary to global photometric alignment proposed in ~\cite{c2f}. An auxiliary data augmentation scheme for global texture alignment is also proposed to reduce the domain gap caused by texture variations.
Experimental results demonstrate that our proposed global manifold alignment and global texture alignment modules make our proposed method more robust and achieve new state-of-the-art performance.

To sum up, this paper has the following new contributions:
\begin{itemize}[noitemsep, nolistsep]
  \item A manifold alignment algorithm is proposed to represent the high-level feature space via dimension reduction and clustering algorithms. To the best of our knowledge, this is the first piece of work that tackles unsupervised domain adaptive semantic segmentation with explicit manifold modeling. All related ablation studies have been conducted for this new module.
  \item Global texture alignment is proposed as a data augmentation scheme for domain adaptive semantic segmentation. It reduces the sensitivity of the trained model with respect to domain-specific textures.
  \item For synthetic source data, our updated method outperforms all previous UDA methods by a large margin, achieving new state-of-the-art performance on both GTA5$\rightarrow$Cityscapes and SYNTHIA$\rightarrow$Cityscapes benchmarks.
  \item We further construct a new medical image domain adaptive semantic segmentation task on the basis of two open-source real-world endoscopic image datasets. Our proposed method also achieves state-of-the-art performance on this new task.
\end{itemize}

\section{Related Work}


\textbf{Photometric Alignment.}
Previous works~\cite{R1, R2, R3} in unsupervised domain adaptation for image classification do not pay attention to image-to-image translation. However, it has been proven that a model trained with source images transferred into the target domain style can significantly improve the final performance in semantic segmentation tasks~\cite{BDL,CAG}. This is perhaps because deep features for semantic segmentation are relatively more sensitive to local information compared to image classification.

In order to achieve image-level photometric alignment, adversarial methods have been widely used in previous work on domain adaptation ~\cite{BDL,dagan1,dagan2,dagan3,dagan4,take_a_look,stuff_things,dagan5}, such as GAN~\cite{gan,cgan} and CycleGAN~\cite{cgan2}. These GAN-based methods can transfer the styles of the images in the target domain to that of the source domain and thus significantly reduce image-level photometric differences ~\cite{dagan3,stuff_things,BDL}. Then, these style transferred source domain images are used to train a segmentation model. Because they are photometrically aligned with the target domain images, the models trained with these style transferred source domain images usually yield better performance compared to the model trained with the source domain images~\cite{stuff_things} only. However, it is also noted that adversarial models are unstable during training. Previous work has shown that image-level adversarial methods generally convert the source domain image-level distribution to the one in the target domain to improve the performance of the domain adapted model~\cite{BDL, stuff_things}. But it is still an open question whether the style transferred images distribution roughly covers the whole target domain image-level distribution or just a small part of it.
The non-adversarial photometric alignment methods for unsupervised semantic segmentation are rare. One latest line of research is the Fourier Domain Adaptation proposed in ~\cite{FDA}. The motivation is that the low-frequency component of an image consists of the major photometric information, and replacing the low-frequency component in a source domain image with its reference image counterpart in the target domain could align the photometric information between different domains. However, the decomposition of frequency components is very sensitive to the image's content, and simply replacing the low-frequency information of an image with that of another image often introduces extra noises and leaves unsatisfactory visual artifacts. According to their experiments, the model's performance trained on the frequency-aligned samples also relies heavily on a multi-band ensemble with multiple models~\cite{FDA}. Unlike the Fourier Domain Adaptation, our proposed method is directly applied to color channels without the frequency decomposition, which provides us with comparable (superior) performance and image quality to its generative (Fourier Transformation-based) counterpart. Moreover, our proposed method only consists of several image-level operations which do not require standalone training and can be used with arbitrary source-target image pairs.


\textbf{Adversarial Methods for Domain Adaptation.}
There are traditional manifold learning methods that model high-dimensional feature spaces before the deep learning era~\cite{isomap, lle, mlle, hlle}, but they are usually computationally costly when transplanted to deep learning applications. Previous work on handling feature spaces in UDA typically adopts adversarial methods~\cite{BDL,fgan}, which do not directly model the feature manifold, but consider features from the source and target domains aligned if they are indistinguishable by a trained discriminator. However, generators trained by adversarial methods are inclined to produce outputs with similar feature distributions~\cite{wgan}. They can surely reduce cross-domain feature distribution discrepancies and make image features agnostic to the input domain. However, it also reduces the diversity of image-level feature distributions from the same domain.
It is difficult to visualize high-dimensional feature distributions resulting from adversarial methods, but we can take style-transferred RGB images generated by adversarial methods as an example. As shown in Figure~\ref{fig:qualitative_gpa}c, all images generated by GAN are dark and smooth regardless of diverse image-level color distributions in the target domain. This phenomenon is called the mode collapse problem and is detrimental to the generalization capability of the domain adapted model in the target domain. Most recent algorithms~\cite{CAG, FDA, fgan} choose to remove adversarial methods from their last stages due to this mode collapse problem.
Our approach differs from adversarial methods in that we model the feature manifold directly by learning a feature manifold from the source domain denoted by a set of representative feature vectors. Then, we propose a pixel feature projection loss that learns to project pixel features from both domains to the source domain feature manifold using these representative feature vectors. Therefore, minimizing the projection errors from both domains benefits domain alignment from a feature-level perspective.

\textbf{Category-Based Methods.} The distribution of different category proportions can be very different between the source domain and the target domain. Existing work \cite{fcns,take_a_look,dagan1,fgan} typically utilizes category labels/predictions to enforce global semantic constraints on category distribution of predicted labels in the target domain.
Similar to their counterparts in image classification~\cite{R1,R2,R3}, some previous works in semantic segmentation (e.g. \cite{CAG} and \cite{stuff_things}) take one step forward to utilize category information: the penultimate image features which are used for generating pseudo-labels in the output layer in the target domain are mapped to their corresponding counterparts in the source domain. Another concurrent work~\cite{zhang2021prototypical} proposes to learn category prototypes online and correct pseudo labels according to the distance measurements between pixels features and those learned category prototypes, which is an improved version of~\cite{CAG}. However, category feature centroids used in ~\cite{CAG, zhang2021prototypical}, or instance features used in ~\cite{stuff_things} only serve as anchors for category-based feature adaptation. The margins between different categories are not explicitly enlarged. This is a problematic alignment strategy because category centroids close to each other in the source domain are still difficult to separate in the target domain. Therefore, we propose a method that differs from theirs in two major perspectives: first, a category-oriented triplet loss is proposed for the source domain to impose a soft constraint that regularizes the category centers for different categories. This approach actively makes inter-category distances between different categories in the high-level feature space larger than intra-category distances of a certain category by a specified margin; secondly, we enforce the predictions on augmented target domain images to be consistent with the pseudo-labels generated by the segmentation model of the corresponding non-augmented images. This is essentially a self-supervision based consistency regularization method and the design philosophy is based on the fact that the supervision signal in the target domain is weak due to the lack of confident pseudo-labels.


\section{Method}
\subsection{Algorithm Pipeline}
\begin{figure*}[ht]
  \centering
  \includegraphics[width=\linewidth]{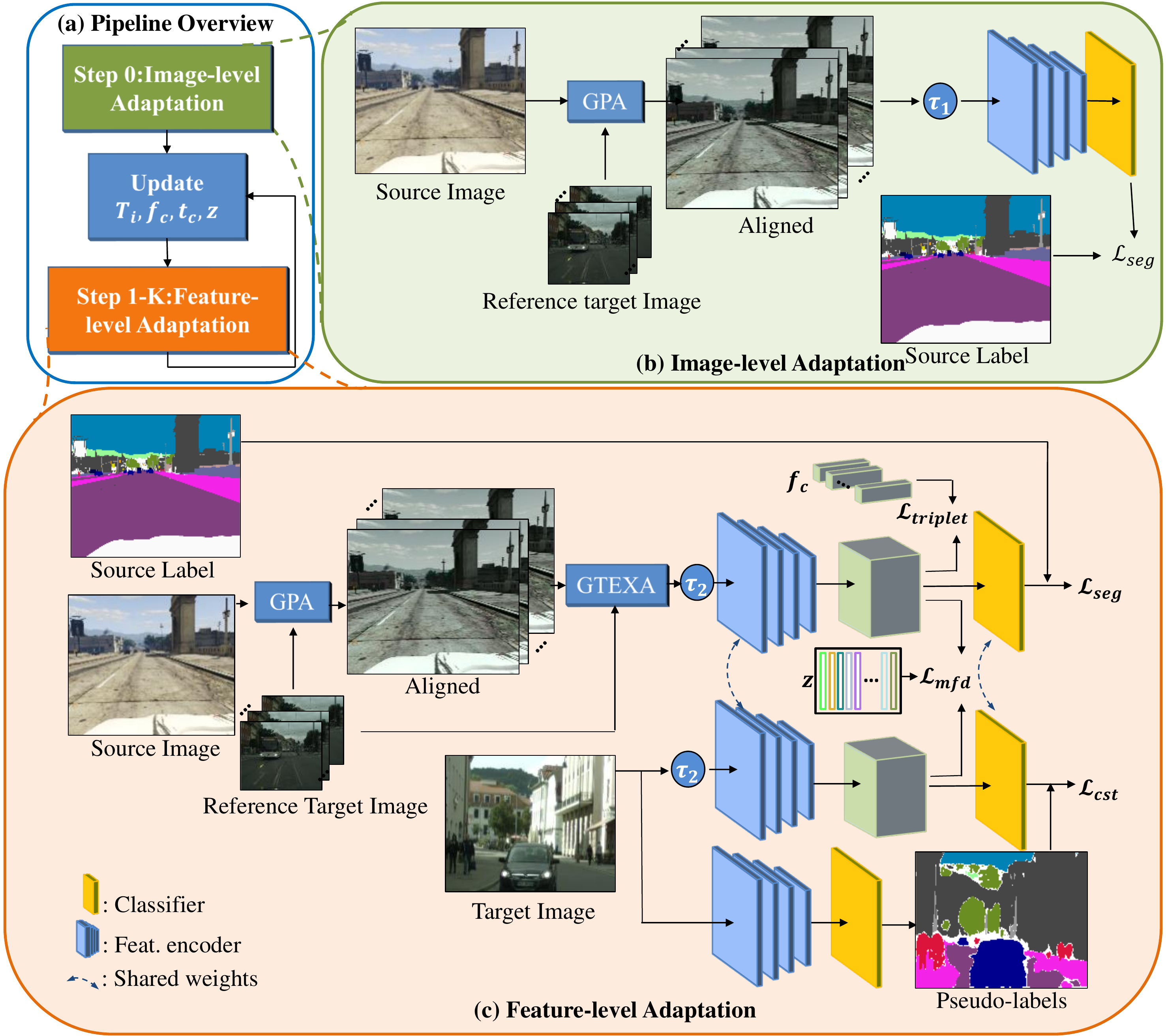} 
  \caption{(a) The pipeline consists of 1 image-level adaptation stage and K feature-level adaptation stages. (b) At first image-level adaptation is implemented using the global photometric alignment operation. (c) Then the obtained model $\mathcal{F}_{i}$ is used to compute pseudo-labels, manifold atoms $\boldsymbol{z}$, category centers $\boldsymbol{f}_{c}$, category thresholds $t_{c}$, as well as initialize the segmentation model for the subsequent feature-level adaptation stages in an iterative self-supervised manner.}
  \label{fig:pipeline}
\end{figure*}

The underlying philosophy of our proposed pipeline is straightforward: first, we exploit the photometric differences in the two domains to coarsely adapt the source domain images with the target domain images to minimize the image-level domain shifts, and the high-frequency distribution from the target domain is also randomly transfered into the source domain image.; then, we perform feature-level adaptation by aligning pixel features from both domains with the feature manifold generated by the coarsely adapted model regardless of its categories; finally, we impose soft constraints on inter-class center distances and intra-class feature variations to regularize category-level feature distributions.  Overview of the pipeline is presented in Figure~\ref{fig:pipeline} and is illustrated as follows.


\textbf{Settings.} Suppose the labeled source domain dataset be $\mathbb D^{s}=\{( \mathbf I^{s}_m, \mathbf Y^{s}_m)\}_{m=1}^{N^s_{I}}$ where $\mathbf I^{s}_m$ is a source image, $\mathbf Y^{s}_m$ is the pixel-level annotation of $\mathbf I^{s}_m$, and $N^s_{I}$ is the number of source images in the source domain dataset. The target domain dataset $\mathbb D^{u}$ contains a large number of unlabeled images $\mathbb D^{u}=\{ \mathbf I^{u}_n\}_{n=1}^{N^{u}_{I}}$. We assume the shape of all images is $h\times w\times 3$, and the number of target classes to be segmented is $M_c$. Hence, we have $\mathbf Y^{s}_m\in \{1,2,\cdots,M_c\}^{h\times w}$. The purpose is to learn a semantic segmentation model for the target domain.


\textbf{Step 0: Image-level Adaptation.} Given a source domain image $\boldsymbol{I}^{s}_{m}$ in the training batch and a randomly selected target domain reference image $\boldsymbol{I}^{u}_{n}$, $\boldsymbol{I}^{s}_{m}$ and $\boldsymbol{I}^{u}_{n}$ are converted into Lab color space as $(\boldsymbol{L}^{s}_{m}, \boldsymbol{a}^{s}_{m}, \boldsymbol{b}^{s}_{m})$ and $(\boldsymbol{L}^{u}_{n}, \boldsymbol{a}^{u}_{n}, \boldsymbol{b}^{u}_{n})$ by our proposed GPA module respectively. The histogram mapping function $f_{match}(\cdot)$ is then used to process both $\boldsymbol{a}^{s}_{m}$ and $\boldsymbol{b}^{s}_{m}$ channels, and gamma correction function $f_{gamma}(\cdot)$ is applied to $\boldsymbol{L}^{s}_{m}$ to form $(f_{gamma}(\boldsymbol{L}^{s}_{m}), f_{match}(\boldsymbol{a}^{s}_{m}), f_{match}(\boldsymbol{b}^{s}_{m}))$. After the mappings, the image is then converted back to RGB space to generate the aligned image $\boldsymbol{\widetilde{I}}^{s}_{m}$. All these randomly generated adapted images are used to construct the adapted source domain training set $\mathbb{\widetilde{D}}^{s}=\{( \boldsymbol{ \widetilde{I}}^{s}_m, \mathbf Y^{s}_m)\}_{m=1}^{N^s}$ for each training epoch. Then, a stochastic function $\tau_1(\cdot)$ is applied to the source domain training set $\mathbb{\widetilde{D}}^{s}$ to produce an augmented version. A segmentation model $\mathcal{F}_0$ is then trained based on the augmented style-transferred source domain images $\tau_1(\mathbb{\widetilde{D}}^{s})$ with the cross-entropy loss $\mathcal{L}_{seg}$.


\textbf{Step 1: Feature-level Adaptation.} The aforementioned image-level adaptation only diminishes the image-level domain shifts between the source and target domains. But image-level adaptation operations do not guarantee the adaptation of high-level features because image components such as textures are not altered by image-level photometric operations, and still impact the high-level features. Therefore, we further modify a random subset of the photometrically aligned images, and make their texture-related high frequency components follow the corresponding distributions in the target domain. Let $\boldsymbol{ \widetilde{I}}^{s}_m$ be a photometrically aligned image, whose texture components are further updated. The resulting image $\boldsymbol{ \overline{I}}^{s}_m$ is the actual input to the segmentation model in this step. We also introduce a global manifold alignment module to tackle the feature-level domain shifts. Before training a new segmentation model, we learn a representation of the feature manifold in the source domain offline. We first apply the initial model $\mathcal{F}_0$ to all source domain images to obtain their feature maps and prediction probability maps. Correctly classified feature vectors from these feature maps are randomly sampled to form matrix $\mathcal{X}$. Then both PCA and K-Means clustering are applied to $\mathcal{X}$ to learn a feature manifold represented with a set of cluster centers $\boldsymbol{z}$ in a dimension reduced feature subspace. When a new segmentation model is trained, features from both the source and target domains are projected onto this manifold, and the projection error, $\mathcal{L}_{mfd}$, is minimized.

In addition to cross-entropy loss $\mathcal{L}_{seg}$ and manifold projection error $\mathcal{L}_{mfd}$, two loss functions for category-level feature distribution regularization are also adopted in the training process. Category center $\boldsymbol{f}_c$ for every category $c$ is calculated as the $L_2$ normalized mean of all pixel features from category $c$ in the source domain. One of the two loss functions is a category-oriented triplet loss $\mathcal{L}_{triplet}$ defined over the image style transferred source domain dataset $\mathbb{\widetilde{D}}^{s}$ to enlarge the inter-category distances and minimize intra-category variances.
In the target domain, the pseudo-label at a certain pixel location and its associated confidence are defined according to the prediction probability maps produced from the initial segmentation model $\mathcal{F}_0$. Pseudo-labels with confidence higher than an adaptive threshold are considered valid samples, and are used to define a target domain consistency loss $\mathcal{L}_{cst}$ to regularize category-level feature distributions in the target domain. The remaining pixels are left out during back-propagation.


We fine-tune the segmentation model $\mathcal{F}_0$ for $U$ iterations by minimizing $\mathcal{L}_{seg}+\mathcal{L}_{mfd}+\mathcal{L}_{triplet}+\mathcal{L}_{cst}$ to produce a new segmentation model $\mathcal{F}_1$ for the current step.

\begin{figure}[ht]
\centering
\includegraphics[width=0.6\linewidth, trim=60 410 580 60, clip]{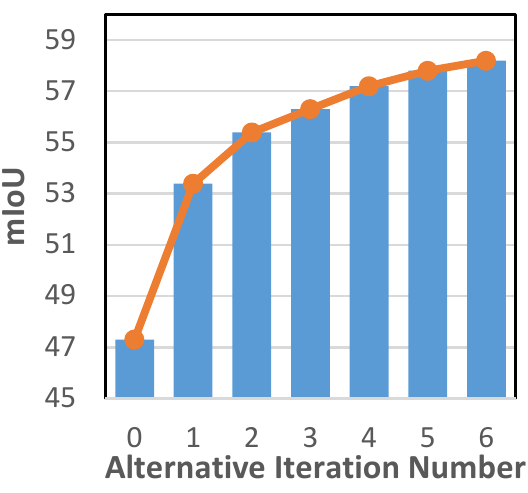} 
\caption{Iterative self-supervised training further improves the segmentation performance.}
\label{fig:iter_chart}
\end{figure}

\textbf{Step 2 to K: Iterative Self-Supervised Training.} Model $\mathcal{F}_1$ trained in Step 1 can be further improved with iterative steps similar to Step 1. Such an iterative approach is called self-supervised training and is widely adopted in the area of unsupervised domain adaptive semantic segmentation~\cite{BDL,CAG,FDA,stuff_things}. The same Step 1 is performed, but the pre-trained model $\mathcal{F}_0$ is replaced with $\mathcal{F}_{i-1}$. And model $\mathcal{F}_{i-1}$ is also used to update manifold atoms $\boldsymbol{z}$, pseudo-labels and category centers $\boldsymbol{f}_c$. This process is repeated for $K-1$ times. Refined pseudo-labels generated by models from each stage further improve the segmentation performance and reduce the domain gap (Figure~\ref{fig:iter_chart}). But erroneous pseudo-labels also accumulate false supervision signals and limit the magnitude of performance improvement. Our proposed image-to-feature pipeline is shown in Figure~\ref{fig:pipeline}.

\subsection{Global Photometric Alignment}
Since global domain shifts are mostly related to low-level image attributes, global photometric alignment is proposed in our work to transfer low-level image attributes of the target domain to source domain images. It is observed that the spatial lightness distribution of an image can be very complicated in different scenarios. It is also important to note that directly operating on RGB channels would cause severe artifacts and fake colors. In contrast, the spatial color distribution of the $a$ and $b$ color channels always have similar bell-shaped histograms. Therefore, we approach lightness and color with different treatments: we perform classic histogram matching~\cite{DIP} between the source domain image and the target domain reference image only on color channels $a$ and $b$ to avoid introducing artifacts commonly seen in histogram matching results.  

\begin{figure}[ht]
  \centering
  \includegraphics[width=\linewidth]{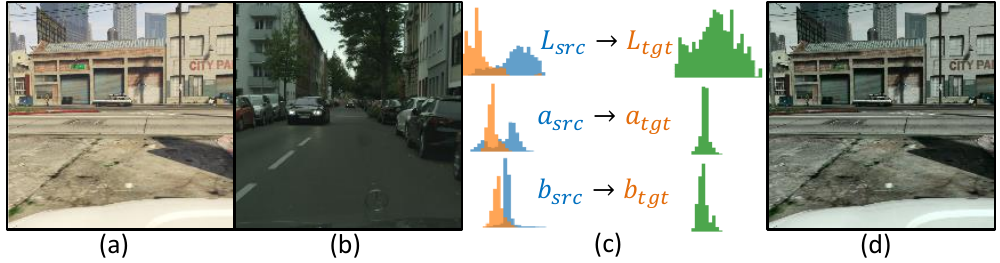}
  \caption{(a)Input source domain image and (b) a randomly chosen target domain image is aligned in (c) Lab channels to generate (d) aligned image.}
  \label{fig:aug}
\end{figure}

\textbf{Lightness Gamma Correction.} On the other hand, the $L$ channel is much more sophisticated under different circumstances. This is because light interacts with the 3D structure of a scene in a complicated manner. Simple histogram matching function results in large areas of overexposure and fake structures. Thus, instead of using histogram matching for every histogram bin to prescribe strict mapping, we choose to constrain the mean value of the lightness channel in the source domain image and make it equal to the mean value of the target domain reference image. Because mean-variance policy might make the pixel value smaller than 0 or larger than 1, we choose the power-law function, which is also widely used in gamma correction. But the difference between our proposed method and the classic gamma correction is that our coefficients for the power-law function are not pre-defined by users. They are automatically calculated with given source-target image pairs. Specifically, the power-law function can be written as $f_{gamma}(L)=L^{\gamma}$, where $L$ is the normalized lightness value from 0 to 1 at each pixel location. $\gamma=1$ when it is an identical transformation.
The mean value constraint can then be written as
\begin{equation}
\sum_{L}f_{gamma}(L)h^s_m(L) = \sum_{L}L^{\gamma}h^s_m(L) = \sum_{L}Lh^u_n(L),
\end{equation}
where $h^s_m$ is the lightness histogram of a source image $\boldsymbol{I}^{s}_{m}$, and $h^u_n$ is the lightness histogram of a target reference image $\boldsymbol{I}^{u}_{n}$.
In practice, we introduce a regularization term $\beta$ to prevent $\gamma$ from deviating too much away from 1. Thus, $\gamma$ can be solved numerically in the following nonlinear optimization,
\begin{equation}
\label{equ:histgamma}
\gamma* = \arg \min_\gamma \left( \sum_{L}L^{\gamma}h^s_m(L) - \sum_{L}Lh^u_n(L) \right)^2+\beta(\gamma-1)^2.
\end{equation}
This optimization problem is a simple convex optimization with only one variable $\gamma$, and can be easily solved with few steps of gradient descent. Source-target image pairs are generated randomly on the fly during training epochs because our proposed GPA is highly efficient and does not require training in comparison to GAN-based methods~\cite{stuff_things, BDL}. The process of the proposed GPA module is illustrated in Figure~\ref{fig:aug}.

\subsection{Global Texture Alignment}
As discussed in previous work~\cite{texture_bias}, CNN-based models are sensitive to high-frequency information. We observe that synthetic images have different and often stronger high-frequency information in comparison to real-world images, which jeopardizes the generalization performance of our model in the target domain. Although the proposed GPA module maintains the diversity of the source domain dataset, it modifies the photometric properties of an image instead of the high-frequency texture. To alleviate this problem, a global texture alignment module is proposed as an auxiliary data augmentation scheme. The idea is straightforward: we modify the high frequency components of a random subset of the source domain images to make their distribution in each image more consistent with that of the corresponding reference image, which is sampled from the target domain. The process is illustrated in Figure~\ref{fig:pipeline}. This data augmentation scheme teaches the segmentation model to ignore texture information and focus on structural information.

To be specific, a bilateral filter $f_{bilateral}(\cdot)|_{d, \sigma_c, \sigma_s}$ is applied to a source domain image $\boldsymbol{\widetilde{I}}^{s}$ to generate the filtered image $\boldsymbol{\overline{I}}^{s}=f_{bilateral}(\boldsymbol{\widetilde{I}}^{s})|_{d, \sigma_c, \sigma_s}$. We use the bilateral filter to preserve image structures and modify the texture component only. In order to determine the parameters ($d$, $\sigma_c$ and $\sigma_s$) of the bilateral filter, we quantify the distribution of high-frequency image components, and ensure that $\boldsymbol{\overline{I}}^{s}$ and its target domain reference image $\boldsymbol{I}^{u}$ have similar distributions of high-frequency components. We convert both $\boldsymbol{\overline{I}}^{s}$ and $\boldsymbol{I}^{u}$ to grayscale images and apply the Laplacian operator $f_{Lap}$ to obtain their high-frequency components $H^s=f_{Lap}(\boldsymbol{\overline{I}})$ and $H^u=f_{Lap}(\boldsymbol{I}^{u})$, respectively. Let $h(H^s)$ and $h(H^u)$ be the respective histogram of $H^s$ and $H^u$, and represent their distribution of high-frequency components. To align $h(H^s)$ and $h(H^u)$, the parameters of the bilateral filter are determined by solving the following optimization problem,
\begin{equation}
\label{equ:histgamma}
d*, \sigma_c*, \sigma_s* = \arg \min_{d, \sigma_c, \sigma_s} KL(\sum_s h(H^s), \sum_u h(H^u))).
\end{equation}
By applying the bilateral filter with optimized parameters, the KL divergence between the distributions of the high-frequency components of $\boldsymbol{\widetilde{I}}^{s}$ and $\boldsymbol{I}^{u}$ can be significantly reduced. Note that $d$, $\sigma_c$ and $\sigma_s$ are fixed once optimized. To introduce stochasticity, each source domain image has 50\% chance to be bilaterally filtered before being fed to the segmentation model. We find that adding this data augmentation scheme in the image-level adaptation step would damage the final performance, and therefore, only use it as an additional source domain data augmentation scheme during the feature-level adaptation steps.

\subsection{Training Loss}
The only training loss during image-level adaptation step is the segmentation cross-entropy loss. The overall loss function we use during feature-level adaptation steps consists of four parts: the cross-entropy segmentation loss, the global manifold alignment loss, the category-oriented triplet loss, and the target domain consistency regularization loss.

\begin{figure}[ht]
  \centering
  \includegraphics[width=0.85\linewidth]{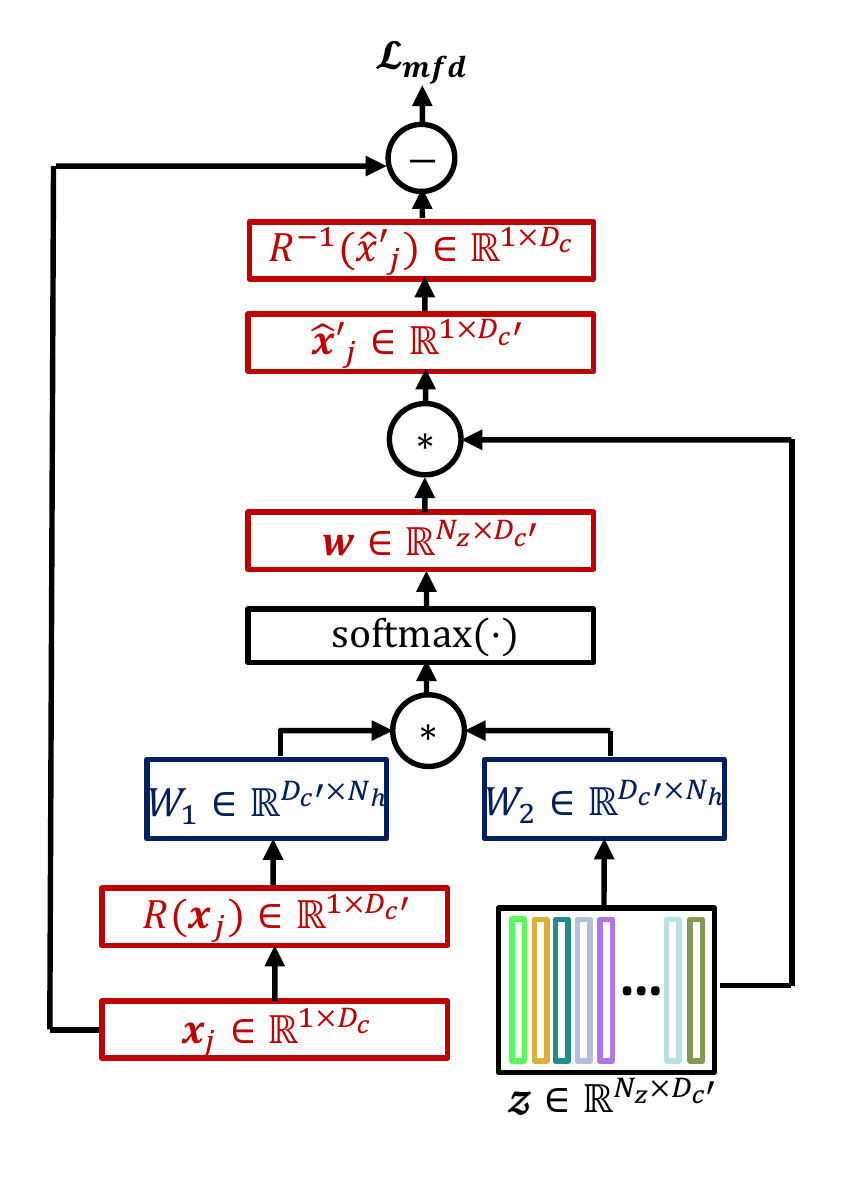} 
  \caption{By minimizing the projection error of source/target domain features onto the manifold, our proposed manifold loss mitigates the discrepancies between source domain feature distribution and target domain feature distribution.}
  \label{fig:manifold_pipeline}
\end{figure}

\textbf{Global Manifold Alignment.}
Methods such as Locally Linear Embedding (LLE) and Isomap are commonly used to depict manifolds, but they are too computationally costly for gradient backpropagation based training. Here we use the K-means algorithm to simplify the computation. As LLE uses a piecewise linear model to approximate a high dimensional feature manifold, K-means can be considered as a piecewise constant approximation of the manifold. Every centroid obtained by K-means is a constant approximation of a local region. By approximating the manifold with a set of representative feature vectors, we can further align features from the source and target domains.

In order to acquire feature representations, first we need to apply the segmentation model obtained in the previous step $\mathcal{F}_{i-1}$ to each source image $\boldsymbol{I}^{s}_{m}$ to compute the feature map of the second last layer $\boldsymbol{X}^{s}_{m}$ and the final prediction probability map $\boldsymbol{P}^{s}_{m}$. The feature vector and prediction probability at a given pixel location $j$ is denoted as $\boldsymbol{x}^{s}_{j}$ and $\boldsymbol{p}^{s}_{j}$, respectively. The true category label at location $j$ is denoted as $\boldsymbol{y}^{s}_{j}$. Next, the predicted probabilities ($\boldsymbol{p}^{s}_{j}$) are compared with true category labels ($\boldsymbol{y}^{s}_{j}$), and the correctly classified feature vectors are randomly sampled to form the source domain sample matrix $\mathcal{X}\in \mathbb{R}^{N_p\times D_{c}}$, where $N_p$ is the total number of sampled feature vectors and $D_{c}$ is the dimensionality of each feature vector.


Then, principal component analysis (PCA) is applied to $\mathcal{X}$ to keep around 90\% of the total explained ratio of energy and obtain the dimension reduced version $R(\mathcal{X})\in \mathbb{R}^{N_p\times D_{c'}}$, where $D_{c'} << D_{c}$. Afterwards, the classic K-Means clustering algorithm is applied to $R(\mathcal{X})$ to find the representative locations on the feature manifold. These locations are denoted as $\boldsymbol{z} \in \mathbb{R}^{N_z \times D_{c'}}$, which are essentially the atom vectors of the source domain feature manifold. Any pixel feature from the source domain ($\boldsymbol{x}^{s}_{j}$) or the target domain ($\boldsymbol{x}^{u}_{j}$) can be projected to the subspace spanned by the atoms in $\boldsymbol{z}$, and the projection is represented as  $\boldsymbol{\hat{x}}'_{j}=\boldsymbol{w}^T\boldsymbol{z}$ (we omit the superscript for simplicity). The projection mapping $\boldsymbol{w}$ is applied here for two reasons: first, although .Let $R^{-1}$ be the reconstruction operator of PCA. The projection error $||R^{-1}(\boldsymbol{\hat{x}}'_{j})-\boldsymbol{x}_{j}||^2$ is considered as the deviation from the source domain manifold and is part of the projection error loss $\mathcal{L}_{mfd}$.

The motivation of our proposed global manifold alignment is straightforward: minimizing the source domain projection error makes the feature manifold smoother, and minimizing the target domain projection error decreases the distance (i.e. improves the alignment) between feature distributions of the source and target domains respectively. Specifically, we adopt an attention mechanism to calculate the linear coefficients of atom vectors. The manifold projection error $\mathcal{L}_{mfd}$ and reconstructed feature vector $\boldsymbol{\hat{x}}'_{j}$ can be computed using the following equations,
\begin{equation}
\label{equ:manifold_loss}
\begin{aligned}
\mathcal{L}_{mfd} &= \sum_j||R^{-1}(\boldsymbol{\hat{x}}'_{j})-\boldsymbol{x}_{j}||^2 \\
\boldsymbol{\hat{x}}'_{j}&=\boldsymbol{w}^T\boldsymbol{z} \\
\boldsymbol{w}^T&=\mbox{softmax}\left( \frac{(R(\boldsymbol{x}_{j}) \boldsymbol{W}_1^T)(\boldsymbol{W}_2\boldsymbol{z}^T)}{\sqrt{N_z}}\right),
\end{aligned}
\end{equation}
where $R(\boldsymbol{x}_{j})$ is the $j$-th row of $R(\mathcal{X})$, both $\boldsymbol{W}_1\in \mathbb{R}^{N_h\times D_{c'}}$ and $\boldsymbol{W}_2\in \mathbb{R}^{N_h\times D_{c'}}$ are trainable linear matrices respectively.   They are introduced to further lower the memory overhead of the attention mechanism. They also project the manifold and all features to a lower dimensional space, and two distinct projection matrices enable better alignment between the projected manifold and features. $N_h$ is a hyperparameter representing the number of hidden neurons, $\boldsymbol{w}\in \mathbb{R}^{N_z}$ is the vector of atom coefficients. The details to calculate the manifold projection loss is illustrated in Figure~\ref{fig:manifold_pipeline}.
Although the global manifold loss is defined for the global alignment of features, it cannot be adopted in the image adaptation stage because it relies on a pre-trained model to provide manifold atoms $\boldsymbol{z}$.

\textbf{Category-oriented Triplet Loss.} Although the aforementioned GPA and GMA modules could learn domain-invariant features to some extent, the losses used in previous training do not explicitly control the category-wise feature distribution, and some category-sensitive domain shifts are overlooked. Pixel features from different categories are naturally distributed unevenly, and some category centers are close to each other. To tackle this issue, we propose a category-oriented triplet loss that aims to push the category-wise features further closer to the corresponding category centers the pixel belongs to and further away from other category centers. Note that category centers are intentionally introduced to make the calculation of category-oriented triplet loss practical. If we use the traditional triplet loss without category centers, we need to store pairwise distances among all pixels with a tremendous GPU memory overload. Therefore, the category center $\boldsymbol{f}_c$ of category $c$ is calculated as follows,
\begin{equation}
\boldsymbol{f}_c = G(\frac{1}{N_c} \sum_s \sum_j  1\left( \boldsymbol{y}^{s}_{j}=c \right)\boldsymbol{x}^{s}_{j}),
\end{equation}
where $\boldsymbol{x}^{s}_{j}$ refers to the pixel-wise features in the penultimate feature map, and $\boldsymbol{y}^{s}_{j}$ be the ground truth pixel-wise labels of a source domain image at pixel location $(j)$. $N_c$ refers to the total number of pixels in category $c$, $s$ refers to the source domain image index, and $G(\cdot)$ is an $L_2$ normalization function. Note that it is crucial to use the $L_2$ normalization $G(\cdot)$ to keep the category centers on the unit sphere and avoid scaling issues among stages. The category centers are updated after the training, allowing the centers to become further and further from each other on the sphere surface.

Our category-oriented triplet loss is formulated as follows,
\begin{equation}
\label{equ:triplet}
\begin{aligned}
\mathcal{L}_{triplet} =& \frac{1}{N_{s}} \sum_s \sum_j \max_{c,c\neq y_j^s} \max ( \left\| G(\boldsymbol{x}^{s}_{j})-\boldsymbol{f}_{y_j^s} \right\| \\ & - \left\| G(\boldsymbol{x}^{s}_{j})-\boldsymbol{f}_{c} \right\| +\alpha, 0),
\end{aligned}
\end{equation}
where $N_{s}$ is the total number of pixels in all images, and $\alpha$ is a prescribed margin. The loss will be zero if every feature $\boldsymbol{x}^{s}_{j}$ is at least $\alpha$ closer to its corresponding category center $y_j^s$ than other category centers. Because triplet loss is focused on hard samples and only reliable category labels for hard samples in the source domain, the proposed category-oriented triplet loss is only applied to the source domain images.

\begin{figure}[ht]
  \centering
  \includegraphics[width=\linewidth]{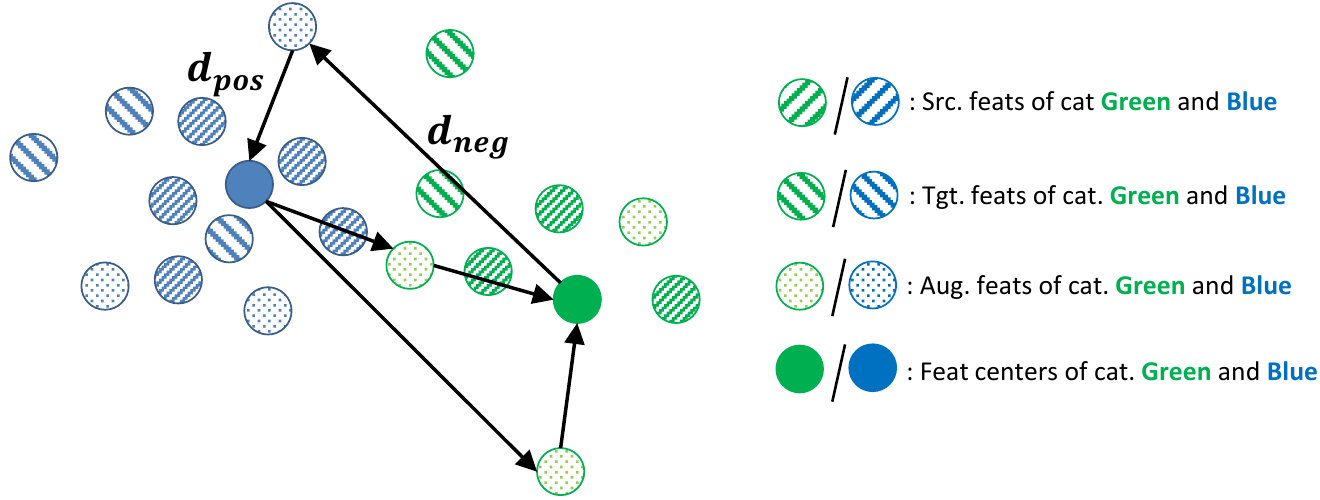} 
  \caption{Our proposed category-oriented triplet loss exploits hard samples and further enlarge category margins. $d_{pos}$ and $d_{neg}$ represent the distance of positive and negative pairs respectively. }
  \label{fig:triplet}
\end{figure}

The working principles of our proposed category-oriented triplet loss are illustrated in Figure~\ref{fig:triplet}. In cooperation with the proposed global photometric alignment and data augmentation in the source domain, our proposed triplet loss can exploit hard samples in the source domain that have been coarsely aligned to the target domain and further improve the generalization capability of the trained model. The proposed category triplet loss can be considered complementary to the cross-entropy loss and the manifold projection loss.

\textbf{Target Domain Consistency Regularization.} The category-wise features are regularized by our proposed category-oriented triplet loss in the source domain, where the annotated ground truth labels are available. However, the supervision signal is weak in the target domain where there is no labeled data provided. Consistency regularization is an important component of many recent state-of-the-art self-supervised learning algorithms, which utilizes unlabeled data by relying on the assumption that the model should output similar predictions when fed perturbed versions of the same image~\cite{consistency, fixmatch}.
Motivated by this, we propose a target domain consistency regularization method shown in Figure~\ref{fig:pipeline} to perform category-level feature distribution regularization in the target domain.

In the target domain, the pseudo-label at a certain pixel location is defined as the category corresponding to the largest component of the probability vector produced from the segmentation model $\mathcal{F}_{i-1}$ trained in the previous step, and the largest component of the probability vector itself defines the confidence of the pseudo label. We further pre-define a pair of probability threshold $P_{h}$ and percentage threshold $p$ for all categories. $p$ is a constant value but leads to a category-specific probability threshold $P_{s, c}$, meaning $p\%$ pixels in the category $c$ have confidence above threshold $P_{s, c}$. Thus the final confidence threshold for category $c$ is $t_c=\min(P_{h}, P_{s, c})$, and any pseudo-labels with a confidence higher than $t_c$ in category $c$ are considered valid samples.

Our proposed target domain consistency regularization is straightforward: given a target domain image $\boldsymbol{I}^{u}_{n}$, with the trained segmentation model $\mathcal{F}_{i-1}$, we extract the pseudo-label $\hat{\boldsymbol{y}}^{u}_{n}$ by forwarding $\boldsymbol{I}^{u}_{n}$ to $\mathcal{F}_{i-1}$ followed by applying the $\arg\max(.)$ function to its output; and the corresponding pixel prediction is converted to a hard label vector $\mathds{1}_{[c=\hat{\boldsymbol{y}}^{u}_{n,j}]}$; then, a stochastic function $\tau_2(\cdot)$ is applied to $\boldsymbol{I}^{u}_{n}$ to obtain a perturbed version $\boldsymbol{\widetilde{I}}^{u}_{n}$; after that, we forward $\boldsymbol{\widetilde{I}}^{u}_{n}$ to $\mathcal{F}_{i}$ to obtain the prediction $\boldsymbol{\widetilde{P}}_{n}^{u}$ on the perturbed image; finally, the prediction $\boldsymbol{\widetilde{P}}_{n}^{u}$ is forced to be consistent with $\hat{\boldsymbol{y}}^{u}_{n}$  by using a cross entropy loss function at pixel locations whose largest class probability is above the previously defined category-level confidence threshold $t_c$. Note that the perturbed target domain image generated via the stochastic function $\tau_2(\cdot)$ makes prediction harder. Thus more samples in the target domain can be converted into hard samples, but the generation of pseudo-labels is unaffected. In this way, category-level feature distributions in the target domain are regularized under the supervision of valid pseudo-labels.
The overall formula is defined as follows,
\begin{equation}
\label{equ:histgamma}
\begin{aligned}
\mathcal{L}_{cst} =& \sum_j \bm{1}(\max(\mathcal{F}_{i-1}(\boldsymbol{I}^{u}_{n})|_j)\geq t_c)\\
&\mbox{CELoss}(\mathds{1}_{[c=\hat{\boldsymbol{y}}^{u}_{n,j}]}, \boldsymbol{\widetilde{P}}^{u}_{n,j}), \\
\hat{\boldsymbol{y}}^{u}_{n,j} =& \arg\max(\mathcal{F}_{i-1}(\boldsymbol{I}^{u}_{n})|_j),\\
\boldsymbol{\widetilde{P}}^{u}_{n,j} =& \mathcal{F}_{i}(\boldsymbol{\widetilde{I}}^{u}_{n})|_j.
\end{aligned}
\end{equation}
It is essential to use the trained model $\mathcal{F}_{i-1}$ rather than model $\mathcal{F}_i$ to generate pseudo-labels. This is because $\mathcal{F}_i$ is still being trained and unstable. Fluctuating pseudo-labels generated by $\mathcal{F}_i$ would be catastrophic to the training process. Experimental results illustrate that this consistency regularization method is simple yet efficient. It strengthens the supervision signal in the target domain and improves the final performance.

\begin{table*}[ht]
  \centering
  \caption{Performance comparison with state-of-the-art methods on the GTA5$\rightarrow$Cityscapes task. Results with image-level adaptation only and our whole image-to-feature pipeline are also presented. The best performing result is marked in \textbf{bold}}
  \resizebox{0.95\linewidth}{!}{
  \begin{tabular}{l l l l l l l l l l l l l l l l l l l l l l l}
    \hline
    & & \rotatebox{90}{road} & \rotatebox{90}{sidewalk} & \rotatebox{90}{building} & \rotatebox{90}{wall} & \rotatebox{90}{fence} & \rotatebox{90}{pole} & \rotatebox{90}{light} & \rotatebox{90}{sign} & \rotatebox{90}{vege}. & \rotatebox{90}{terrace} & \rotatebox{90}{sky} & \rotatebox{90}{person} & \rotatebox{90}{rider} & \rotatebox{90}{car} & \rotatebox{90}{truck} & \rotatebox{90}{bus} & \rotatebox{90}{train} & \rotatebox{90}{motor} & \rotatebox{90}{bike} & mIoU \\
    \hline
    \multirow{8}{*}{DeeplabV2} & BDL~\cite{BDL} & 91.0 & 44.7 & 84.2 & 34.6 & 27.6 & 30.2 & 36.0 & 36.0 & 85.0 & \textbf{43.6} & 83.0 & 58.6 & 31.6 & 83.3 & 35.3 & 49.7 & 3.3 & 28.8 & 35.6 & 48.5 \\
     & IDA~\cite{intra_domain} & 90.6 & 36.1 & 82.6 & 29.5 & 21.3 & 27.6 & 31.4 & 23.1 & 85.2 & 39.3 & 80.2 & 59.3 & 29.4 & 86.4 & 33.6 & 53.9 & 0.0 & 32.7 & 37.6 & 46.3 \\
     & DTST~\cite{stuff_things} & 90.6 & 44.7 & 84.8 & 34.3 & 28.7 & 31.6 & 35.0 & 37.6 & 84.7 & 43.3 & 85.3 & 57.0 & 31.5 & 83.8 & 42.6 & 48.5 & 1.9 & 30.4 & 39.0 & 49.2\\
     & FGGAN~\cite{fgan} & 91.0 & 50.6 & \textbf{86.0} & \textbf{43.4} & \textbf{29.8} & 36.8 & 43.4 & 25.0 & \textbf{86.8} & 38.3 & 87.4 & 64.0 & \textbf{38.0} & 85.2 & 31.6 & 46.1 & 6.5 & 25.4 & 37.1 & 50.1\\
     & FDA~\cite{FDA} & \textbf{92.5} & \textbf{53.3} & 82.3 & 26.5 & 27.6 & 36.4 & 40.5 & 38.8 & 82.2 & 39.8 & 78.0 & 62.6 & 34.4 & 84.9 & 34.1 & 53.1 & \textbf{16.8} & 27.7 & 46.4 & 50.4 \\
     & image adap. (ours) & 84.6 & 37.4 & 81.0 & 25.6 & 12.9 & 35.7 & 33.8 & 16.5 & 83.5 & 31.2 & 82.7 & 64.8 & 35.7 & 85.3 & 30.0 & 31.9 & 8.0 & 25.7 & 32.2 & 44.1\\
    & I2F~\cite{c2f} & 89.8 & 46.0 & 85.8 & 32.5 & 22.3 & \textbf{41.0} & 43.9 & 28.9 & 86.4 & 31.0 & \textbf{89.4} & \textbf{65.6} & 36.9 & \textbf{87.9} & 42.4 & 54.4 & 6.5 & 38.9 & \textbf{56.2} & 51.9\\
    & I2F all (ours) & 90.8 & 48.7 & 85.2 & 30.6 & 28.0 & 33.3 & \textbf{46.4} & \textbf{40.0} & 85.6 & 39.1 & 88.1 & 61.8 & 35.0 & 86.7 & \textbf{46.3} & \textbf{55.6} & 11.6 & \textbf{44.7} & 54.3 & \textbf{53.3}\\
     \hline
    \multirow{6}{*}{DeeplabV3/+} & CAG~\cite{CAG} & 90.4 & 51.6 & 83.8 & 34.2 & 27.8 & \textbf{38.4} & 25.3 & \textbf{48.4} & 85.4 & 38.2 & 78.1 & 58.6 & 34.6 & 84.7 & 21.9 & 42.7 & 41.1 & 29.3 & 37.2 & 50.2 \\
    & WCBT~\cite{pitfall} & 89.4 & 50.1 & 83.9 & \textbf{35.9} & 27.0 & 32.4 & 38.6 & 37.5 & 84.5 & 39.6 & 85.7 & 61.6 & 33.7 & 82.2 & 36.0 & 50.4 & 0.3 & 33.6 & 32.1 & 49.2 \\
    & image adapt. (ours) & 83.9 & 37.5 & 82.7 & 28.7 & 18.9 & 35.3 & 41.3 & 31.1 & 85.2 & 29.5 & 86.6 & 62.8 & 30.9 & 82.4 & 23.0 & 39.3 & 33.0 & 26.0 & 39.7 & 47.3\\
    & I2F~\cite{c2f} & 92.5 & 58.3 & 86.5 & 27.4 & 28.8 & 38.1 & \textbf{46.7} & 42.5 & 85.4 & 38.4 & \textbf{91.8} & 66.4 & 37.0 & 87.8 & 40.7 & 52.4 & 44.6 & 41.7 & 59.0 & 56.1\\
    & I2F all (ours) & \textbf{92.6} & \textbf{59.1} & \textbf{87.0} & 33.3 & \textbf{32.2} & 37.2 & 46.3 & 47.5 & \textbf{86.2} & \textbf{40.2} & 90.7 & \textbf{67.6} & \textbf{41.1} & \textbf{88.1} & \textbf{48.7} & \textbf{53.7} & \textbf{47.0} & \textbf{47.4} & \textbf{60.6} & \textbf{58.2}\\
    \hline
  \end{tabular}}
  \label{tab:gta2city}
\end{table*}

\begin{table*}[ht]
  \centering
  \caption{Performance comparison with state-of-the-art methods on the Synthia$\rightarrow$Cityscapes task (mIoU: 16-class; mIoU*: 13-class). The best performing result is marked in \textbf{bold}}
  \resizebox{0.95\linewidth}{!}{
  \begin{tabular}{l l l l l l l l l l l l l l l l l l l l l l l}
    \hline
    & & \rotatebox{90}{road} & \rotatebox{90}{sidewalk} & \rotatebox{90}{building} & \rotatebox{90}{wall} & \rotatebox{90}{fence} & \rotatebox{90}{pole} & \rotatebox{90}{light} & \rotatebox{90}{sign} & \rotatebox{90}{vege}. & \rotatebox{90}{sky} & \rotatebox{90}{person} & \rotatebox{90}{rider} & \rotatebox{90}{car} & \rotatebox{90}{bus} & \rotatebox{90}{motor} & \rotatebox{90}{bike} & mIoU & mIoU*\\
    \hline
    \multirow{8}{*}{DeeplabV2} & BDL~\cite{BDL} & \textbf{86.0 } & \textbf{46.7 } & 80.3 & - & - & - & 14.1 & 11.6 & 79.2 & 81.3 & 54.1 & 27.9 & 73.7 & 42.2 & 25.7 & 45.3 & - & 51.4 \\
    & IDA~\cite{intra_domain} & 84.3 & 37.7 & 79.5 & 5.3 & 0.4 & 24.9 & 9.2 & 8.4 & 80.0 & 84.1 & 57.2 & 23.0 & 78.0 & 38.1 & 20.3 & 36.5 & 41.7 & 48.9 \\
    & DTST~\cite{stuff_things} & 83.0 & 44.0 & 80.3 & - & - & - & 17.1 & 15.8 & 80.5 & 81.8 & 59.9 & 33.1 & 70.2 & 37.3 & 28.5 & 45.8 & - & 52.1 \\
    & FGGAN~\cite{fgan} & 84.5 & 40.1 & \textbf{83.1} & 4.8 & 0.0 & 34.3 & 20.1 & 27.2 & \textbf{84.8} & 84.0 & 53.5 & 22.6 & \textbf{85.4 } & 43.7 & 26.8 & 27.8 & 45.2 & 52.5 \\
    & FDA~\cite{FDA}  & 79.3 & 35.0 & 73.2 & - & - & - & 19.9 & 24.0 & 61.7 & 82.6 & 61.4 & 31.1 & 83.9 & 40.8 & \textbf{38.4 } & 51.1 & - & 52.5 \\
    & image adapt. (ours) & 76.4 & 28.8 & 71.6 & 7.7 & 0.5 & 31.0 & 13.8 & 27.8 & 69.3 & 70.0 & 59.7 & 26.4 & 75.7 & 29.9 & 22.1 & 25.2 & 39.7 & 45.9 \\
    & I2F~\cite{c2f} & 81.9 & 33.7 & 78.5 & \textbf{11.0} & \textbf{1.9} & \textbf{36.7} & 32.6 & 33.4 & 79.6 & 78.2 & 67.3 & 33.6 & 84.0 & 33.5 & 25.9 & 47.6 & 47.5 & 54.6 \\
    & I2F all (ours) & 84.9 & 44.7 & 82.2 & 9.1 & \textbf{1.9} & 36.2 & \textbf{42.1} & \textbf{40.2} & 83.8 & \textbf{84.2} & \textbf{68.9} & \textbf{35.3} & 83.0 & \textbf{49.8} & 30.1 & \textbf{52.4} & \textbf{51.8} & \textbf{60.1} \\
    \hline
    \multirow{6}{*}{DeeplabV3/+} & CAG (13 classes)~\cite{CAG}  & \textbf{84.8} & 41.7 & \textbf{85.5 } & - & - & - & 13.7 & 23.0 & \textbf{86.5 } & 78.1 & \textbf{66.3 } & 28.1 & 81.8 & 21.8 & 22.9 & 49.0 & - & 52.6 \\
    & CAG (16 classes)~\cite{CAG} & 84.7 & 40.8 & 81.7 & 7.8 & 0.0 & 35.1 & 13.3 & 22.7 & 84.5 & 77.6 & 64.2 & 27.8 & 80.9 & 19.7 & 22.7 & 48.3 & 44.5 & - \\
    & WCBT~\cite{pitfall} & 81.7 & \textbf{43.8} & 80.1 & \textbf{22.3} & 0.5 & 29.4 & \textbf{28.6} & 21.2 & 83.4 & 82.3 & 63.1 & 26.3 & 83.7 & 34.9 & 26.3 & 48.4 & 47.2 & 54.1 \\
    & image adapt. (ours) & 64.0 & 25.7 & 73.9 & 9.6 & 0.8 & 33.3 & 12.3 & 25.9 & 81.6 & 85.5 & 62.4 & 26.2 & 80.6 & 30.9 & 26.8 & 23.8 & 41.5 & 47.7 \\
    & I2F~\cite{c2f} & 75.7 & 30.0 & 81.9 & 11.5 & 2.5 & \textbf{35.3} & 18.0 & \textbf{32.7} & 86.2 & \textbf{90.1} & 65.1 & \textbf{33.2} & 83.3 & 36.5 & 35.3 & \textbf{54.3} & 48.2 & 55.5 \\
    & I2F all (ours) & 79.9 & 34.2 & 83.6 & 11.4 & \textbf{4.8} & 35.3 & 28.0 & 34.8 & \textbf{86.9} & 91.1 & 64.9 & 32.5 & \textbf{86.2} & \textbf{50.1} & \textbf{42.6} & 52.9 & \textbf{51.2 } & \textbf{59.1} \\
    \hline
  \end{tabular}}
  \label{tab:synthia2city}
\end{table*}

\section{Experiments}
\subsection{Datasets and Implementation Details}
For commonly used synthetic datasets, we follow the same evaluation settings as used in \cite{CAG}. Our proposed method are evaluated with datasets \textbf{GTA5}~\cite{gta5}, \textbf{Synthia}\cite{synthia}, and \textbf{Cityscapes}~\cite{cityscapes}. The \textbf{Cityscapes} dataset is the target domain dataset with $2,957$ of size $2048\times1024$ training images and $500$ validation images of the same resolution. \textbf{Cityscapes} has 19 categories of objects in total. The \textbf{GTA5} and \textbf{Synthia} are two source domain datasets of computer generated synthetic images, which contain $24,966$ of size $1914\times1052$ training images and $9400$ of size $1280\times760$ training images respectively. The \textbf{GTA5} dataset shares 19 common categories with the \textbf{Cityscapes} dataset, and all the irrelevant categories are ignored during training. The \textbf{Synthia} dataset shares 16 common categories with the \textbf{Cityscapes} dataset. Some previous work~\cite{BDL,FDA} only train and test on a 13-category subset of the \textbf{Synthia} dataset, or train two models on both subset and the whole set for better performance~\cite{CAG}. Here we follow the practice in \cite{intra_domain, fgan} to train a model only on the whole set and test it on both settings.

In order to evaluate the performance of our proposed method on real-world source images, we construct a new domain adaptive semantic segmentation task Kvasir$\rightarrow$Piccolo on the basis of two open-source datasets \textbf{Hyper-Kvasir}~\cite{kvasir} and \textbf{Piccolo}~\cite{piccolo}. The \textbf{Hyper-Kvasir} dataset is the source dataset and consists of $1000$ wide-band (WL) gastrointestinal images. The image resolution of the \textbf{Hyper-Kvasir} dataset is not fixed and is roughly $625\times530$. The Piccolo dataset is the target dataset. Among the $1302$  narrow-band (NBI) colonoscopy images of size $854\times480$, $1161$ are used as training images and $141$ as validation images. We follow the rule in ~\cite{pitfall} to construct this new task specifically designed for medical images: the images were collected with different modes (NBI vs. WL), different locations (GI tract vs. colonoscopy) and different devices to create a significant domain gap between the source and target domains.

According to Figure~\ref{fig:pipeline}, the photometrically adapted source domain images are used first to train the initial segmentation model $\mathcal{F}_0$ in the image-level adaptation step. Then, the model is trained in an iterative self-supervision manner with $K=6$ and $U=20k$. We compare to previous work in domain adapted semantic segmentation~\cite{CAG,stuff_things} based on self-supervision and set the total number of training iterations to $140k$. As reported in ~\cite{BDL}, the best performance is achieved when $P_h=0.9$ and $p=10$ for pseudo-labels, and we follow this setting in our experiment. The regularization term $\beta$ used by GPA module in (\ref{equ:histgamma}) is set to $0.01$. For the proposed global texture alignment (GTEXA) module, $d=5$, $\sigma_c=75$ and $\sigma_s=25$ are the optimized parameters of the bilateral filter for the GTA5$\rightarrow$Cityscapes and Synthia$\rightarrow$Cityscapes tasks, the KL divergence is reduced roughly from 0.16 to 0.10 and from 0.43 to 0.07, respectively. Because images from both \textbf{Hyper-Kvasir} and \textbf{Piccolo} are real images, they have quite similar distributions of high-frequency components. We use a gentle bilateral filter with $d=5$, $\sigma_c=10$ and $\sigma_s=25$. The KL divergence stays at 0.10 before and after the bilateral filter is applied. For the proposed GMA module, $D_{c'}$ is set to keep roughly $90\%$ explained ratio of the energy of $\boldsymbol{x}_{j}$, which is $D_{c'}=32$ for DeeplabV3+ and $D_{c'}=256$ for DeeplabV2, compared to $D_{c}=256$ for DeeplabV3+ model and $D_{c}=2048$ for DeeplabV2 respectively. The K-Means is used because it is the simplest clustering algorithm to validate our motivation. The number of clusters centers are set to $N_z=64$ with hidden neuron dimension $N_h=32$. Note that a larger $N_z$ or a more advanced clustering method like KSVD might improve the performance. Still, it is not pragmatic because of the memory consumption or the computational complexity. We adopt the standard color-jittering as the stochastic function $\tau_1(\cdot)$ in both source, and target domains as in \cite{fgan} in the image-level adaptation stages. We utilize standard color-jittering, elastic deformation~\cite{UNet}, and standard random blurring in the feature-level adaptation stages. Elastic deformation is used to mimic the differences between shapes in different domains, and random blurring is used to simulate the resolution differences. We conduct different settings for $\tau_1(\cdot)$ and $\tau_2(\cdot)$ because we observed that both elastic deformation and random blurring are strong data augmentations, and using them in the image adaptation stage will distort the distribution of the training data, which undermines the final segmentation performance.

We follow the same experiment settings in ~\cite{c2f}. In addition to the DeeplabV3+(ResNet101)~\cite{deeplab} discussed in ~\cite{c2f}, we also compare our proposed model with another commonly adopted segmentation model DeeplabV2(ResNet101) used by other state-of-the-art studies~\cite{fgan,stuff_things,BDL}. We implement our proposed method with PyTorch~\cite{pytorch}, and deploy our experiments on 4 NVIDIA GeForce 2080Ti GPUs with 1 source domain image and 1 target domain image randomly selected and stored on each GPU for each backpropagation step. The stochastic gradient descent is used during the image-level adaptation with a momentum of $0.9$ and weight decay of $1e−4$. The learning rate is initially set to $5e−4$ and is decreased using the polynomial learning rate policy with a power of $0.9$ during training. For the feature-level adaptation steps, we halve the learning rate to $2.5e−4$ based on the learning rate from the image-level adaptation to fine-tune previously trained models.

\begin{figure*}[ht]
  \centering
  \includegraphics[width=0.95\linewidth]{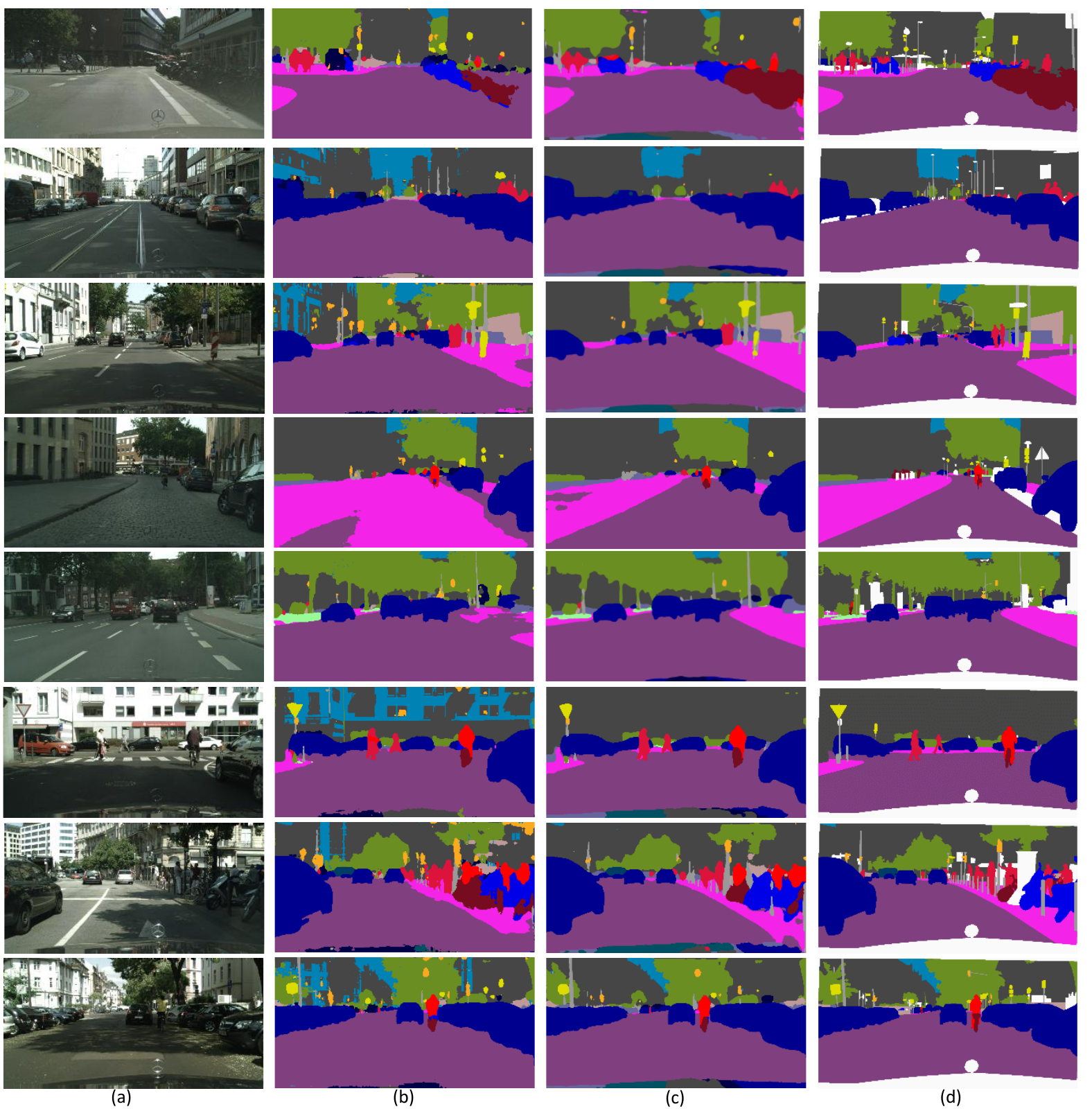} 
  \caption{Qualitative examples of the comparisons between our method and CAG~\cite{CAG} on the GTA5$\rightarrow$Cityscapes task. Specifically, (a) Input images, (b) CAG~\cite{CAG}, (c)Ours, (d) Labels}
  \label{fig:qualitative_piccolo}
\end{figure*}

\begin{figure*}[ht]
  \centering
  \includegraphics[width=0.8\linewidth]{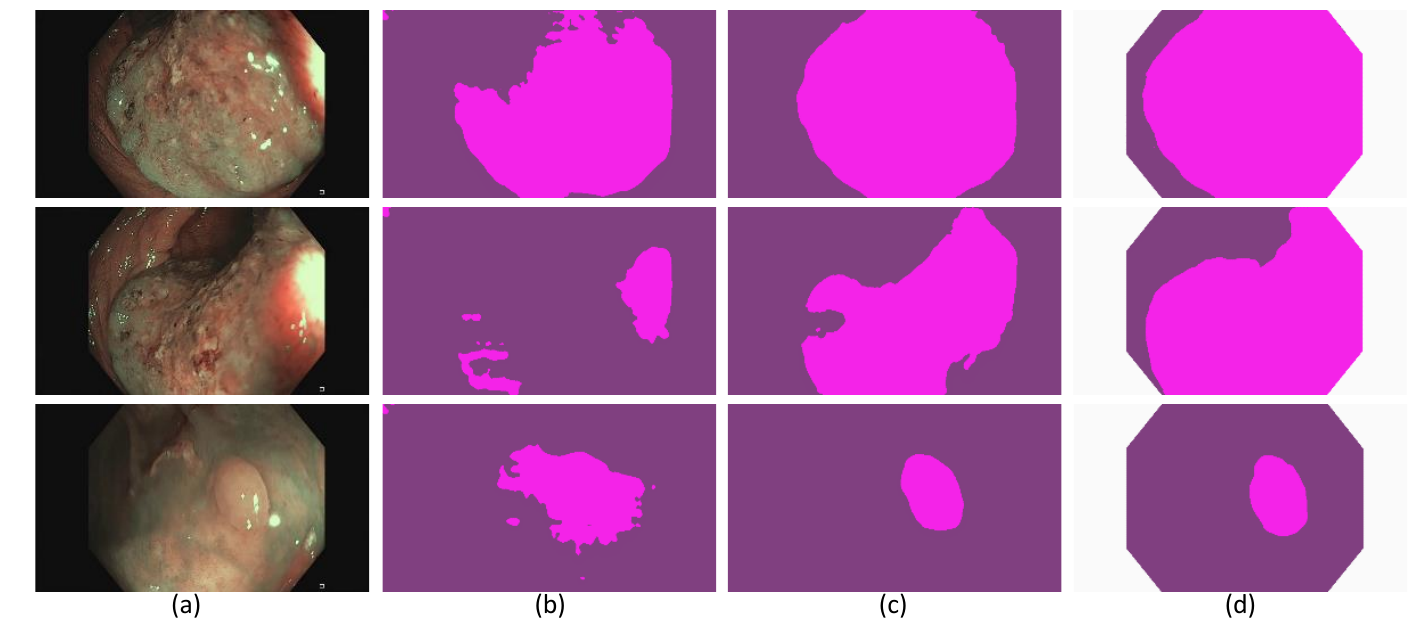} 
  \caption{Qualitative examples of the comparisons between our method and CAG~\cite{CAG} on the Kvasir$\rightarrow$Piccolo task. Specifically, (a) Input images, (b) CAG~\cite{CAG}, (c)Ours, (d) Labels}
  \label{fig:qualitative}
\end{figure*}

\subsection{Comparisons with State-of-the-Art Methods}
In this section, we compare our method against all the existing state-of-the-art methods~\cite{intra_domain,stuff_things,CAG,FDA,fgan,BDL}, on the GTA5$\rightarrow$Cityscapes, Synthia$\rightarrow$Cityscapes and Kvasir$\rightarrow$Piccolo tasks. As shown in Table~\ref{tab:gta2city}, the performance improvement achieved by our proposed method outperforms all previous methods with all different segmentation models. On the GTA5$\rightarrow$Cityscapes task, our model achieves a new state-of-the-art mIoU ($58.2\%$) on DeeplabV3+ and ($53.3\%$) on DeeplabV2, which are $8.0\%$ and $2.9\%$ higher than the previous best result using the same backbone (Table~\ref{tab:gta2city}), respectively. On the Synthia$\rightarrow$Cityscapes task, the performances of our proposed method are $60.1\%$ and $59.1\%$, which are $7.6\%$ and $5.0\%$ higher than that of the previous method, respectively. Note that we do not include the comparison between our work and a concurrent work~\cite{zhang2021prototypical} because three major differences make the comparison unfair, which are presented as follows: (1) much stronger baseline models. \cite{zhang2021prototypical} adopts the domain adapted model from \cite{dagan1} as the baseline model, which generates much stronger performance than ours. For example, our baseline model ($37.6\%$) is $5.8\%$ lower than theirs ($43.4\%$) on the GTA5$\rightarrow$Cityscapes task; (2) much stronger segmentation networks. \cite{zhang2021prototypical} uses a more advanced segmentation network proposed in ~\cite{zheng2019unsupervised, zheng2020unsupervised} instead of the standard DeeplabV2 that we used; (3) much stronger augmentation strategies. RandAug~\cite{Cubuk2020RandaugmentPA} and Cutout~\cite{Devries2017ImprovedRO} are utilized as data augmentations in~\cite{zhang2021prototypical}, which are much superior than our augmentations. On the Kvasir$\rightarrow$Piccolo task, the segmentation performance of our proposed method is $81.5\%$ and $84.2\%$ on DeeplabV2 and DeeplabV3+ respectively, which are $3.7\%$ and $6.0\%$ higher than the previous best results, as shown in Table~\ref{tab:medseg}. We re-implemented three general-purpose state-of-the-art methods~\cite{FDA,fgan,CAG} for this comparison. Our proposed method also significantly outperforms the domain adaptive segmentation algorithm in \cite{pitfall}, which was specifically designed for endoscopic images.

Our method achieves the best performance in many important categories, including `road', `sidewalk', `building', `fence', `vegetation', `terrace', `person', `car', `rider', `truck', `train', `bus', `motor', and `bike'. In particular, our model performs the best when classifying over `road', `sidewalk', `motor', and `bike', even if some of these categories have very similar local appearances. This is because our proposed category-oriented triplet loss maximizes the inter-category distances and minimizes the intra-category distances by exploiting the most difficult samples in the source domain, which improves the generalization capability across different domains. Moreover, our proposed global manifold/texture alignment brings an extra $2.1\%$ improvements in the final performance compared to the experiments in ~\cite{c2f}. This is because global photometric alignment is generally conducted in the image-level inputs, and the features/textures between different domains are still not explicitly aligned. Finally, our proposed target consistency regularization also strengthens the relatively weak supervision signal in the target domain. It improves the segmentation accuracy of categories with large intra-category variances, such as `building' and `sky', by regularizing its feature distribution. According to our qualitative analysis, the previously over-exposed white buildings are easily misclassified as skies but can be corrected by our proposed target consistency regularization.

\begin{figure*}[ht]
\centering
\includegraphics[width=\linewidth]{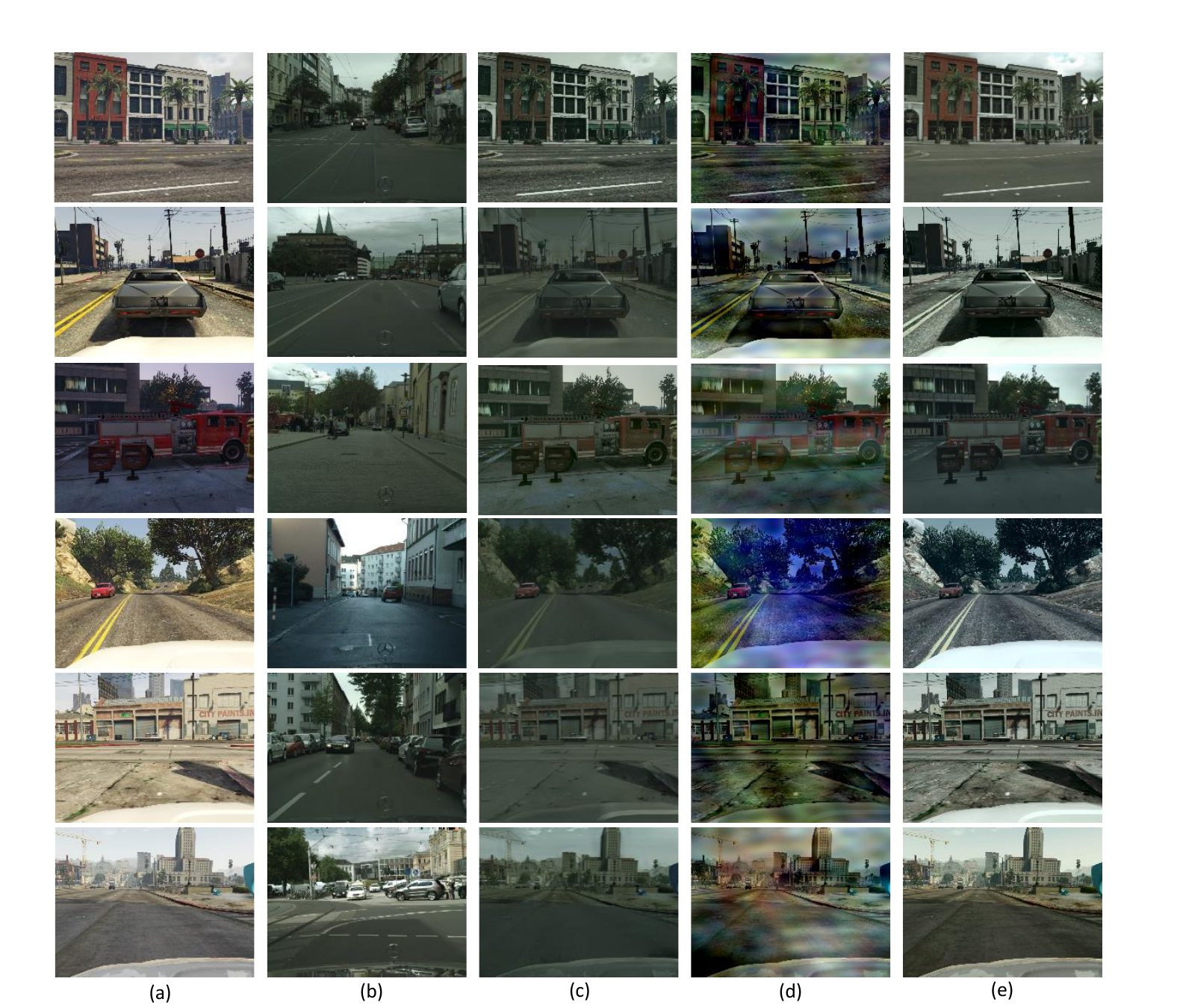} 
\caption{Qualitative analysis on the global photometric alignment (GPA) module. (a) Input images, (b) Reference images, (c) BDL-GAN\cite{BDL}, (d) Fourier Adaptation\cite{FDA}, (e) Global Photometric Alignment.}
\label{fig:qualitative_gpa}
\end{figure*}

One interesting fact is that our proposed method has larger overall performance improvements in the GTA5$\rightarrow$Cityscapes task compared to the Synthia$\rightarrow$Cityscapes task. This is because DeeplabV3+ adopts high-resolution feature maps and lower feature dimensions, which improves the segmentation performance. The Synthia dataset is mostly constituted by large objects, limiting the improvements that a DeeplabV3+ segmentation model can bring.

Another interesting fact is that although our proposed GMA has achieved clear performance improvements in both DeeplabV3+ and DeeplabV2, yet the performance improvement with DeeplabV2 is higher as showin in Table~\ref{tab:gta2city}. This is because the feature dimension for DeeplabV2 is much larger than the feature dimension for DeeplabV3+(2048 vs. 256). This results in a more complicated feature manifold which is difficult to align with simple image-level photometric alignment. By modeling the manifold and minimize the projection error, our proposed GMA can effectively align high-dimensional features. Although DeeplabV3+ is powerful, by cooperating with our proposed GMA module, the performance of our proposed model on the Synthia$\rightarrow$Cityscapes task with DeeplabV2 is even higher than the one with DeeplabV3+.

\begin{table}[h]
  \centering
  \caption{Performance comparison with state-of-the-art methods on the Kvasir$\rightarrow$Piccolo task. Best results are marked in bold.}
  \resizebox{\linewidth}{!}{
  \begin{tabular}{l l l l l l l l l l l l l l l l l l l l l l l}
    \hline
    & & polyps & background & mIoU \\
    \hline
    \multirow{3}{*}{DeeplabV2} & FGGAN~\cite{fgan} & 62.2 & 88.5 & 75.4\\
    & FDA~\cite{FDA} & 66.2 & 89.5 & 77.8\\
     & image adapt.~\cite{c2f} & 54.7 & 84.9 & 69.8\\
    & I2F all (ours) & \textbf{72.0} & \textbf{90.9} & \textbf{81.5}\\
     \hline
    \multirow{4}{*}{DeeplabV3/+} & WCBT~\cite{pitfall} & 56.9 & 86.5 & 76.3\\
    & CAG~\cite{CAG} & 67.0 & 89.4 & 78.2\\
    & image adapt.~\cite{c2f} & 55.8 & 82.0 & 68.9\\
    & I2F all (ours) & \textbf{76.2} & \textbf{92.2} & \textbf{84.2}\\
    \hline
  \end{tabular}}
  \label{tab:medseg}
\end{table}

We further show some of the segmentation examples in Figure~\ref{fig:qualitative} and Figure~\ref{fig:qualitative_piccolo} to qualitatively demonstrate the superiority of our method. Our proposed method generates finer edges and makes fewer mistakes. We also compare the style-transferred images generated by our proposed GPA model with other state-of-the-art style-transfer techniques used by the domain adaptation methods in Figure~\ref{fig:qualitative_gpa}, and our proposed method has better quality and a higher level of diversity.

\subsection{Ablation Studies}
\textbf{Component Analysis.} In most previous work, a source-only model trained on the source domain training set only is often required to serve as initial pseudo label producer. Although we do not use the source-only model during training, we train one to provide a baseline so that it is convenient to verify the primary performance gains from our proposed pipeline. Then, following the experiment settings in~\cite{CAG}, we perform extensive ablative experiments using DeeplabV3+ on the GTA5$\rightarrow$Cityscapes task to verify the effectiveness of each of our proposed component.


As shown in Table~\ref{tab:modules}, the source-only baseline using Deeplab v3+ has a performance of  $37.6\%$  for the GTA5$\rightarrow$Cityscapes task, and our proposed overall pipeline improves the baseline performance by $20.6\%$. Following the same settings in previous state-of-the-art methods~\cite{CAG,intra_domain,fgan}, we further evaluate the impact of each proposed component according to the final performance of our model on the GTA5$\rightarrow$Cityscapes task by removing one component at a time. The results are shown in Table~\ref{tab:modules}.  According to our results, the final performance of the segmentation model has the most deterioration when the global photometric alignment module is removed. This is because the GPA module is critical to the image-level adaptation. Removing it literally removes the first image-level adaptation stage, and thus, the resulting erroneous pseudo-labels are harmful to later stages. Although our proposed global manifold alignment module can align the feature distributions from different domains, the error from unaligned models still accumulates across steps, which is detrimental to the final performance of the model. This also validates the necessity of a image-level adaptation step and the importance of an accurate initial model.

Even though our proposed GPA module serves as an image-level adaptation between domains, the feature-level domain shifts are still not aligned completely. Our proposed GMA module can further improve the feature-level adaptation between the source domain and the target domain, decreasing the model performance by $1.4\%$ when removing the GMA module. In addition, the GTEXA module is designed to modify the high-frequency components of an image with a certain probability. This action makes trained models robust to texture variations and further improves the performance by roughly $0.7\%$. Furthermore, despite its simplicity, our proposed target domain consistency regularization has been proved to be very effective. The main reason for this phenomenon is that there are fewer valid training samples in the target domain than the source domain, and our proposed TCR essentially serves as a data augmentation technique that increases the number of valid training samples in the target domain. It also introduces more hard samples without damaging the pseudo-labels. Therefore, it gives rise to a significant performance gain, decreasing the model performance by $4.8\%$ when removing the TCR module. Our proposed category-oriented triplet loss applied on the source domain also boosts the performance by $3.1\%$ as it exploits hard samples in the source domain and improves generalization capability across different domains.

\begin{table}[h]
  \centering
  \caption{Ablation study of the proposed components on the GTA5$\rightarrow$Cityscapes task. GPA: global photometric alignment, GTEXA: global texture alignment, CTL: category-oriented triplet loss, TCR: target domain consistency regularization.}
  \resizebox{\linewidth}{!}{
  \begin{tabular}{l l l l l l l}
    \hline
    & GPA & GTEXA & GMA & CTL & TCR & mIoU \\
    \hline
    Source only & & & & & & 37.6\\
    Image adapt. & $\surd$ & & & & & 47.3\\
    w/o Alignments & & & & $\surd$ & $\surd$ & 47.5\\
    w/o GPA & & $\surd$ & $\surd$ & $\surd$ & $\surd$ &  51.0\\
    w/o GMA & $\surd$ & $\surd$ & & $\surd$ & $\surd$ & 56.6\\
    w/o CTL & $\surd$ & $\surd$ & $\surd$ & & $\surd$ & 55.1\\
    w/o TCR & $\surd$ & $\surd$ & $\surd$ & $\surd$ & & 53.4 \\
    w/o GTEXA & $\surd$ & & $\surd$ & $\surd$ & $\surd$ & 57.5\\
    all & $\surd$ & $\surd$ & $\surd$ & $\surd$ & $\surd$ & 58.2 \\
    \hline
  \end{tabular}}
  \label{tab:modules}
\end{table}

\textbf{Photometric Alignment.}
There are currently other methods, which can achieve the goal of the image-level adaptation, such as the GAN-based method in \cite{BDL} and the frequency-based method in \cite{FDA}. We substitute our proposed global photometric alignment and global texture alignment with these two methods and retrain our whole pipeline. The result is shown in Table~\ref{tab:ablations}. We also visualize some representative aligned images produced with different methods in Figure~\ref{fig:qualitative_gpa}. Our proposed GPA can generate the aligned image according to a randomly chosen target domain reference image. Simultaneously, the GAN-based model~\cite{BDL} performs deterministically and generates aligned images with a similar style, only covering part of the actual target domain image span. This explains why our proposed model works even better than the pre-trained deep adversarial model. Although the frequency-based method proposed in \cite{FDA} can generate style-transferred images randomly, the concatenation of frequencies usually introduces significant noises during training, which largely limits its final performance.

Based on our observation, gamma correction on Lab channels does not have sufficient adaptation capability, while histogram matching on all three channels results in image artifacts. We use the simple mean-variance of RGB channels as the benchmark, and run a comparison for the image-level adaptation stage. The result in Table~\ref{tab:ablations} shows our hybrid scheme performs the best.

\begin{figure}[ht]
\centering
\includegraphics[width=\linewidth]{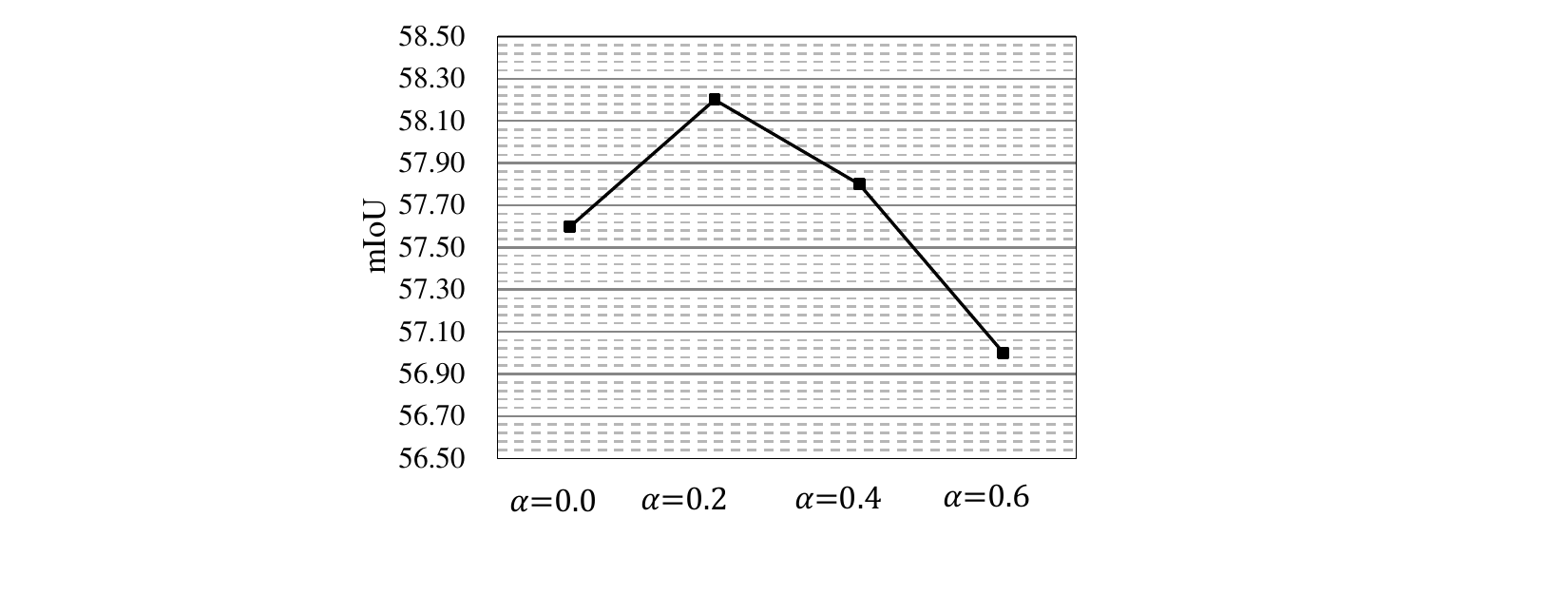} 
\caption{Quantitative analysis on the selection of category margin $\alpha$. Best performance is achived with $\alpha=0.2$.}
\label{fig:hyper_ctl}
\end{figure}

\textbf{Category Triplet Loss.} We only apply the category-oriented triplet loss to the source domain category labels in our proposed method but not the pseudo-labels in the target domain. Although the target domain images with pseudo-labels can be used as supplementary samples when the pseudo-labels are of high confidence, our proposed triplet loss aims to deal with hard samples and pseudo-labels of hard samples in the target domain are not reliable. To verify this, we include pseudo-labels in our category-oriented triplet loss, and the result is shown in Table~\ref{tab:ablations}. We follow the default settings as in \cite{triplet} and set $\alpha=0.2$. But we also tested our proposed category triplet loss with other settings as shown in Figure~\ref{fig:hyper_ctl}, which shows that the best result is achieved when $\alpha=0.2$.

\begin{figure}[ht]
\centering
\includegraphics[width=\linewidth]{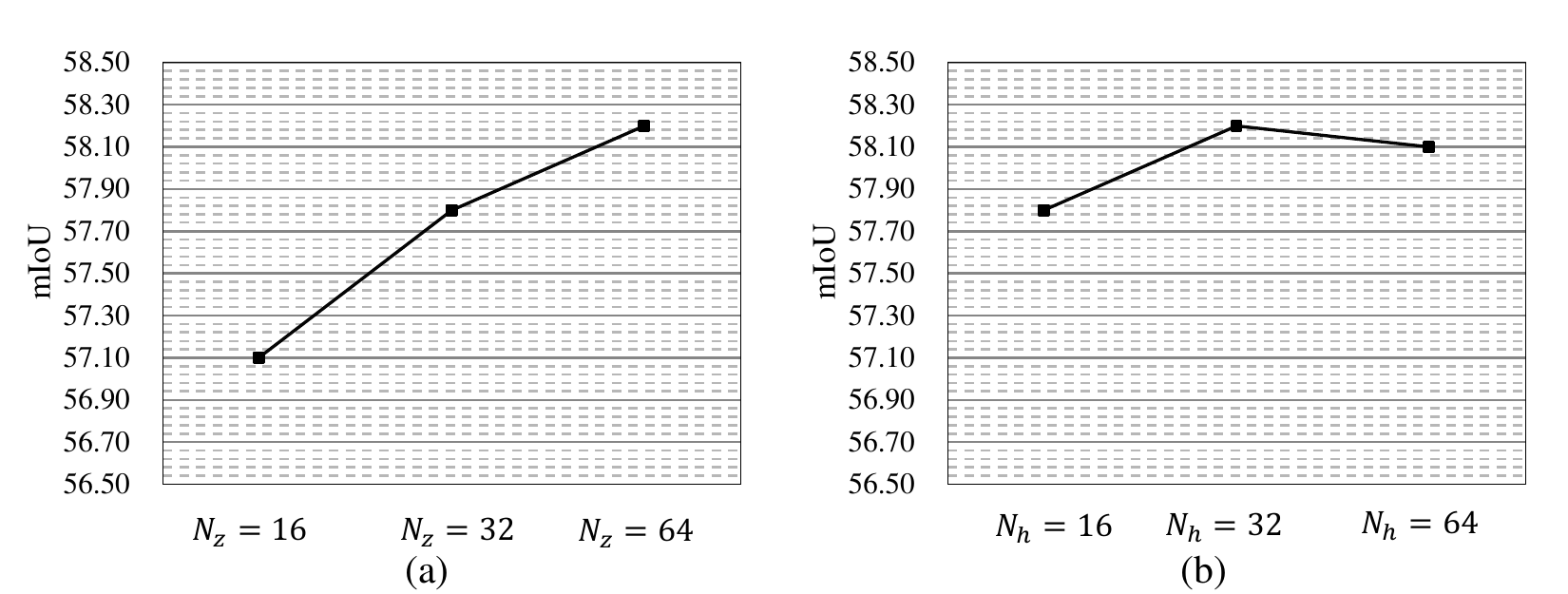} 
\caption{Quantitative analysis on the hyper-parameters of global manifold alignment (GMA) module. (a) higher $N_z$ leads to better performance, (b) mIoU is generally not very sensitive to the choice of $N_h$.}
\label{fig:hyper_mfd}
\end{figure}

\textbf{Manifold Alignment.} In our proposed model, we directly use the PCA+K-Means to model the feature manifold. It shares similar functionality with the adversarial methods used in previous work. But our proposed method suffers little from mode collapse. The mode collapse is easily observed in the style-translated images as in Figure~\ref{fig:qualitative_gpa}. Still, it also exists in the high-level features and undermines the diversity of the training set. To make fair comparisons, we substitute our proposed manifold alignment module with a traditional global discriminator as in~\cite{stuff_things}. According to Tabel~\ref{tab:ablations}, the result illustrates that it performs even worse than the version without a global discriminator, manifesting the superiority of our GMA module. We also conduct extensive experiments to verify the values of the hyperparameters $N_h$ and $N_z$. The experiment results are presented in Figure~\ref{fig:hyper_mfd}. In general, the performance of the model is insensitive to the choice of $N_h$, and larger $N_z$ leads to better performance, the best performance is achieved with settings $N_h=32$ and $N_z=64$. Note that we can not increase atom number $N_z$ to more than $64$ because the calculation of atom weights and the reconstruction of pixel feature $\boldsymbol{x}_{j}$ require a large amount of GPU memory.

\begin{table}[h]
\centering
\caption{Ablation studies of the image adaptation strategy, photometric alignment scheme, and using pseudo-labels for the category-oriented triplet loss on the GTA5$\rightarrow$Cityscapes task.}
\label{tab:ablations}
\resizebox{\linewidth}{!}{
\begin{tabular}{lll}
\hline
Modules & Methods & mIoU \\
\hline
\multirow{3}{*}{Image Aapt.} & Frequency Align~\cite{FDA}. & 54.0 \\
 & BDL-GAN~\cite{BDL} & 55.4 \\
 & Photometric+Texture & 58.2 \\
\hline
\multirow{3}{*}{GPA Scheme} & RGB Mean-Variance & 42.3 \\
& Lab Gamma Correction & 44.5 \\
 & Lab Histogram Match & 43.3 \\
 & Hybrid & 47.3 \\
\hline
\multirow{2}{*}{Pseudo-labels} & Triplet loss with pseudo-labels & 56.1 \\
 & Triplet loss w/o pseudo-labels & 58.2 \\
\hline
\multirow{2}{*}{Manifold Align.} & Adversarial Method &  55.8\\
& Manifold Alignment & 58.2 \\
\hline
\end{tabular}}
\end{table}

\section{Conclusions}
In this paper, we have explored non-adversarial methods in both image-level and feature-level domain adaptation, and proposed a novel unified image-to-feature adaptation pipeline for unsupervised domain adaptive semantic segmentation. During this study, we have found out that for this specific problem, adversarial methods could damage the diversity of feature distributions, and a simple photometric alignment module can achieve better performance. We have also found out that a simple self-supervised consistency loss is capable of regularizing category-level feature distributions in the target domain. The proposed pipeline effectively integrates global image-level and feature-level adaptation and category-level feature distribution regularization. The global texture alignment module also serves as an auxiliary data augmentation scheme for the proposed pipeline.

In particular, we have introduced a novel and efficient global photometric alignment module to adapt source domain images to the target domain. A global texture alignment module has been designed to modify the high-frequency components of images from the source domain and make the trained model robust to domain gaps caused by domain-specific textures. We have also proposed a global manifold alignment module to directly model the distribution of the pixel features from the source domain and align the feature distributions from both domains. To our best knowledge, this is the first piece of work that models the feature manifold directly in unsupervised domain adaptation for semantic segmentation. A category-oriented triplet loss has been devised for the source domain to regularize source domain category centers. A target domain consistency regularization method has also been introduced for the target domain to regularize category-level feature distributions. Extensive experiments have shown that each of our proposed techniques improves the generalization capability of our model significantly. The proposed three modules form a complete adaptation strategy to tackle domain shifts. Integrating them gives rise to a significant improvement over existing state-of-the-art unsupervised domain adaptive semantic segmentation methods, demonstrating that minimizing global and category-level domain shifts simultaneously deserves more attention.

\textbf{Limitations.} Our work still has a few limitations. First of all, the photometric alignment module is isotropic, which means texture information is not altered by our proposed module. Although we have proposed a global texture alignment scheme, it is activated only when the source domain images have stronger or similar high-frequency components in comparison to the target domain images. It deserves more attention to develop a method that can better close the domain gap without hurting the feature diversity of source domain samples. In addition, our scheme for global feature manifold alignment is the first attempt to model the feature manifold directly. However, when we design our scheme, the priority is making manifold alignment compatible with gradient back-propagation based training, but not achieving optimal alignment performance. Nonetheless, it demonstrates the potential of direct feature manifold modeling in domain adaptation tasks. At last, some of our proposed components are only designed for the close-set setting because they are based on the assumption that deep features from the same category should be similar, which is not well suited for open-set tasks where different unseen categories are all labeled as ``unknown". Thus how to extend our algorithm to the open-set scenario remains an open problem.

\ifCLASSOPTIONcaptionsoff
  \newpage
\fi



\bibliographystyle{IEEEtran}
\bibliography{egbib}
\end{document}